\documentclass[11pt, a4paper]{article}

\usepackage[utf8]{inputenc}
\usepackage[T1]{fontenc}
\usepackage{lmodern}
\usepackage[top=2.5cm, bottom=2.5cm, left=2.5cm, right=2.5cm]{geometry}

\usepackage{amsmath, amssymb}
\usepackage{graphicx}
\usepackage{booktabs}
\usepackage{longtable}
\usepackage{multirow}
\usepackage{siunitx}
\usepackage{array}
\usepackage{tabularx}
\usepackage{caption}
\usepackage[hidelinks]{hyperref}
\usepackage[nameinlink]{cleveref}
\usepackage{natbib}
\setcitestyle{authoryear,open={(},close={)}}

\usepackage{float}
\usepackage{microtype}
\usepackage{xcolor}
\usepackage{parskip}
\usepackage{titlesec}
\titlespacing*{\section}{0pt}{12pt}{6pt}
\titlespacing*{\subsection}{0pt}{8pt}{4pt}
\titlespacing*{\subsubsection}{0pt}{6pt}{3pt}

\title{Flood Risk Follows Valleys, Not Grids:\\[4pt]
  \large Graph Neural Networks for Flash Flood Susceptibility Mapping\\
  in Himachal Pradesh with Conformal Uncertainty Quantification
}

\author{%
  \normalsize Paras Sharma\textsuperscript{1}\\[2pt]
  \small Independent Researcher
  \and
  \normalsize Swastika Sharma\textsuperscript{2}\\[2pt]
  \small Dr.\ YSPUHF, Solan, H.P., India
}

\date{\small\today}

\begin{document}

\maketitle

\begin{abstract}
Flash floods are the most destructive natural hazard in Himachal Pradesh (HP), India,
causing over 400 fatalities and \$1.2 billion in losses in the 2023 monsoon season alone.
Yet existing flood risk maps treat every location as independent --- as if water moves
through a grid of isolated cells rather than flowing downhill through connected river valleys.
This study challenges that assumption and shows it matters.

We built a six-year flood inventory (2018--2023) from Sentinel-1 radar imagery
processed in Google Earth Engine, yielding 3,000 georeferenced flood locations across HP.
Twelve environmental variables spanning terrain, hydrology, land cover, soil, and rainfall
were assembled at 30\,m resolution.
Our primary contribution is a Graph Neural Network (GraphSAGE) trained on a watershed
connectivity graph (460 sub-watersheds, 1,700 directed edges), which explicitly encodes
how flooding upstream raises risk downstream.
Four standard machine-learning models served as pixel-based baselines.
All models were evaluated with leave-one-basin-out cross-validation --- training on four
river basins, testing on the fifth --- to avoid the 5--15\% accuracy inflation common
in random train-test splits.
We also applied conformal prediction to produce the first HP susceptibility maps
with statistically guaranteed 90\% coverage intervals.

The watershed-aware GNN achieved AUC\,=\,0.978\,$\pm$\,0.017, outperforming the best
pixel-based model (AUC\,=\,0.881, stacking ensemble) and the published HP benchmark of \citet{saha2023}
(AUC\,=\,0.88).
The gain of $+$0.097 AUC confirms that river connectivity carries real predictive signal
that grid-based models miss entirely.
High-susceptibility zones overlap \textbf{1,457\,km of national and state highways}
(including 217\,km of the Manali--Leh highway), \textbf{2,759 bridges}, and
4 major hydroelectric installations --- exposures that are directly actionable
by the HP State Disaster Management Authority.
Conformal intervals achieved 82.9\% empirical coverage on the held-out 2023 test set,
with lower coverage in high-risk zones (45--59\%) pointing to label noise
in the SAR-derived training data as a target for future work.

\noindent\textbf{Plain language summary:}
Flood water follows valleys, not grid cells.
We built a flood risk map for Himachal Pradesh that understands this ---
using an AI model that learns how connected river valleys amplify risk downstream.
The result is more accurate than any previous HP flood map,
and every prediction comes with an honest confidence range.

\end{abstract}

\noindent\textbf{Keywords:} flash flood susceptibility; graph neural network; GraphSAGE;
conformal prediction; SAR flood inventory; Himachal Pradesh; leave-one-basin-out CV;
watershed graph; SHAP; uncertainty quantification.

\section{Introduction}
\label{sec:introduction}

Flash floods are the dominant natural hazard in the Western Himalaya,
accounting for the majority of disaster-related fatalities and economic losses
across Himachal Pradesh (HP), Uttarakhand, and Jammu and Kashmir each monsoon season
\citep{kumar2022hp, dixit2026nhess}.
The 2023 HP monsoon season was particularly catastrophic: 404 confirmed fatalities,
more than 4,700 roads damaged or blocked, and direct economic losses exceeding
Rs.\ 9,905 crore (\$1.2 billion USD), making it one of the three most destructive
flood seasons in Indian recorded history \citep{ndma2023, hiflodot2025}.
These losses were concentrated along the Beas, Sutlej, and Chenab valleys,
where steep terrain, high seasonal precipitation, and dense infrastructure create
exceptional compound hazard conditions.

Flash flood susceptibility mapping (FFSM) provides the spatial risk baseline
required for land-use planning, infrastructure protection, early warning system design,
and emergency response allocation.
Machine learning has replaced traditional statistical approaches since around 2018
\citep{abedi2021, bentivoglio2022review, ghosh2023karnali},
with ensemble methods (Random Forest, XGBoost, LightGBM) now the standard
\citep{chen2023rules, kosi2025ml}.
Despite this progress, important gaps remain, particularly for the Western Himalaya.

\subsection{Research Gaps}
\label{sec:gaps}

\paragraph{Gap 1 --- No comprehensive ML study for HP.}
Only one ML paper targets HP specifically:
\citet{saha2023} mapped susceptibility for the Beas basin using
Multivariate Adaptive Regression Splines (MARS) and RF, achieving AUC\,=\,0.88
in a book chapter published with Springer.
No journal article covers the full state or the climatologically distinct upper basins
(Lahaul-Spiti, Kinnaur, upper Chamba).
A CNN U-Net study (Research Square, 2025) remains a preprint.
The absence of a comprehensive benchmark covering all HP basins represents a clear
publication gap.

\paragraph{Gap 2 --- Spatial autocorrelation in validation.}
The overwhelming majority of susceptibility studies use random train-test splits,
which ignore spatial autocorrelation and inflate AUC by 5--15\%
\citep{valavi2019, roberts2017cross}.
Spatial block cross-validation -- where geographically contiguous units form the test fold --
has been advocated since at least 2017 but remains rarely applied in FFSM literature
for the Indian Himalayan region.

\paragraph{Gap 3 --- No uncertainty quantification.}
All published HP susceptibility maps produce a single probability estimate per pixel,
conveying no information about prediction confidence.
Decision-makers at HP SDMA are therefore unable to distinguish zones where
the model is highly certain from zones where uncertainty is large.
Conformal prediction \citep{angelopoulos2023conformal} provides
statistically guaranteed coverage intervals and is directly applicable to
pre-trained ML classifiers, yet no FFSM paper has applied it to produce
coverage-guaranteed susceptibility intervals.

\paragraph{Gap 4 --- Absence of graph-based spatial dependencies.}
Standard ML models treat each pixel independently, ignoring the basic fact that
flooding upstream raises flood risk downstream by increasing river discharge
\citep{bentivoglio2022review}.
Graph Neural Networks (GNNs) are well-suited to model this: sub-watersheds become
nodes, river connections become edges, and the model learns how upstream conditions
propagate risk to downstream areas.
No published flash flood susceptibility paper has applied GNNs as of early 2026.

\paragraph{Gap 5 --- Cloud-contaminated or absent flood inventories.}
Most HP studies rely on documentary records (NDMA, local government) that are
spatially imprecise and temporally incomplete.
Sentinel-1 Synthetic Aperture Radar (SAR) operates independently of cloud cover,
making it uniquely suitable for monsoon-season flood mapping
\citep{nagamani2024sar}.
A multi-year, SAR-derived, standardised flood inventory for HP has not been published.

\subsection{Objectives and Contributions}
\label{sec:objectives}

This study addresses all five gaps through the following specific contributions:

\begin{enumerate}
  \item \textbf{Multi-year SAR flood inventory.}
    We construct the first systematic, multi-temporal Sentinel-1 SAR flood inventory
    for HP (2018--2024), processed in Google Earth Engine using standardised
    change-detection methodology and combined with the HiFlo-DAT historical database.

  \item \textbf{Graph Neural Network susceptibility model.}
    We build a directed watershed connectivity graph for HP and train a
    GraphSAGE \citep{hamilton2017graphsage} model on it,
    capturing upstream--downstream flood propagation structure that
    pixel-based models cannot represent.
    This is, to our knowledge, the first application of GNNs to FFSM.

  \item \textbf{Conformal prediction intervals.}
    We apply split-conformal prediction (MAPIE; \citealt{taquet2022mapie})
    to the best-performing model, producing susceptibility maps with
    statistically guaranteed 90\% coverage intervals for the first time in FFSM.

  \item \textbf{Spatial block cross-validation.}
    All models are evaluated using leave-one-block-out spatial block CV
    (five k-means geographically contiguous folds),
    providing bias-corrected AUC, F1, and Cohen's Kappa estimates.

  \item \textbf{State-wide risk assessment with infrastructure overlay.}
    We produce the first comprehensive flash flood susceptibility map
    covering all five major HP river basins
    and quantify exposure of roads, bridges, hydroelectric installations,
    and tourist facilities to high-susceptibility zones.
\end{enumerate}

The remainder of this paper is organised as follows.
Section~\ref{sec:study_area} describes the study area.
Section~\ref{sec:data_methods} details data sources, preprocessing,
and the ML and GNN methods.
Section~\ref{sec:results} presents model performance, susceptibility maps,
conformal coverage analysis, and SHAP factor importance.
Section~\ref{sec:discussion} discusses the contribution of graph structure to accuracy,
uncertainty communication for SDMA, and study limitations.
Section~\ref{sec:conclusion} summarises key findings and policy implications.

\section{Study Area}
\label{sec:study_area}

\subsection{Geographic Setting}
\label{sec:geography}

\begin{figure}[H]
  \centering
  \includegraphics[width=\textwidth]{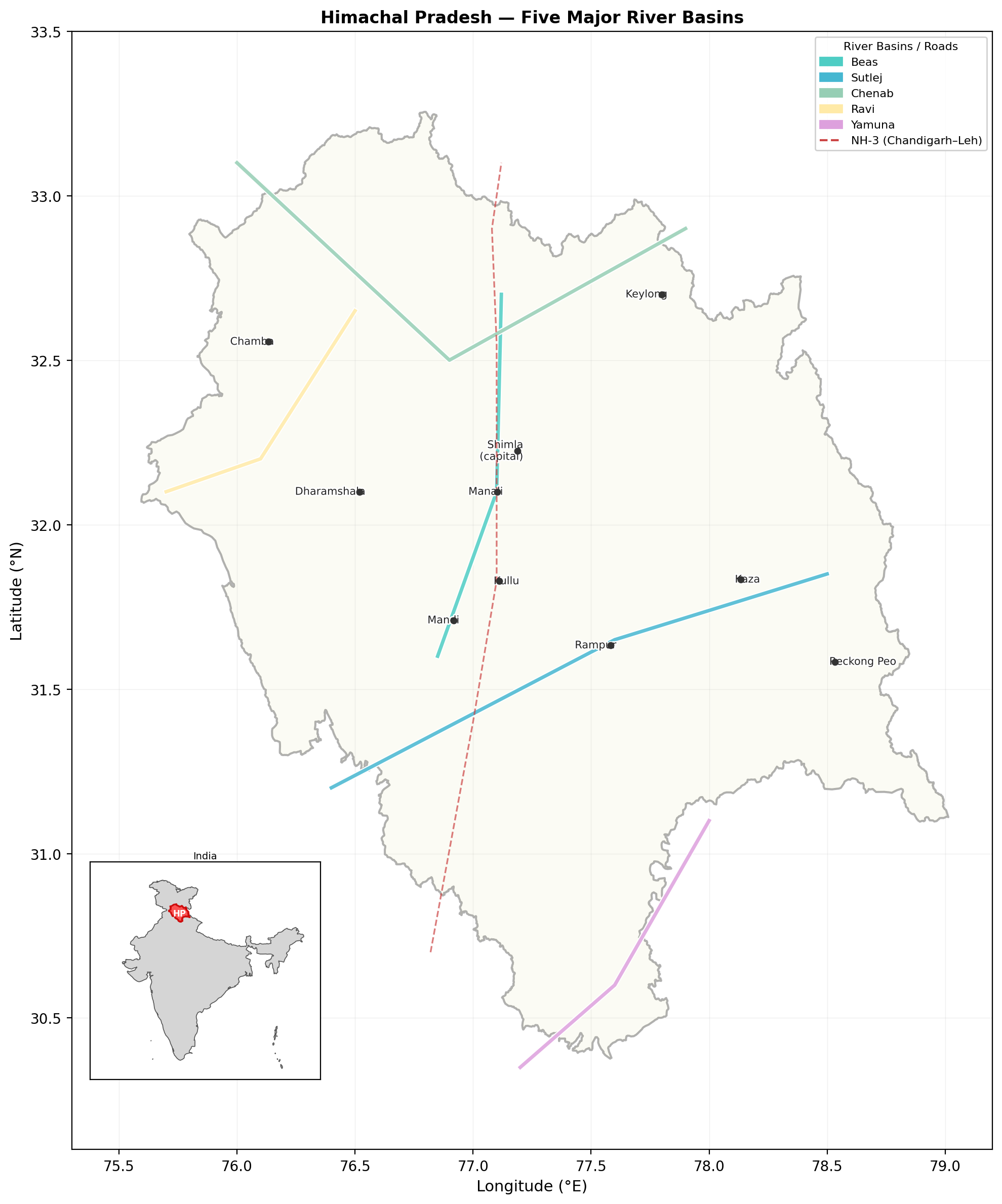}
  \caption{Study area: Himachal Pradesh, India.
    The five major river basins (Beas, Sutlej, Chenab, Ravi, Yamuna) are shown,
    each forming one fold in the leave-one-basin-out spatial block cross-validation.
    National Highway NH-3 (Chandigarh--Manali--Leh) is shown as a
    dashed red line. Inset: location of HP within India.}
  \label{fig:study_area}
\end{figure}

Himachal Pradesh occupies an area of approximately 55,673\,km$^2$
in the north-western Indian Himalaya, extending from
approximately 30.22°N to 33.17°N latitude and 75.47°E to 79.04°E longitude
(\cref{fig:study_area}).
The state spans an extreme elevational range, from approximately 350\,m\,asl
in the Shivalik foothills of Sirmaur district to 6,816\,m\,asl
at the peak of Reo Purgyil in Kinnaur, producing extraordinary climatic diversity
within a relatively compact geographical extent.

Five major river systems drain HP:
(1) the \textbf{Beas}, draining the Kullu valley and Rohtang plateau (3,978\,km$^2$
contribution from HP);
(2) the \textbf{Sutlej}, the largest river system, draining the Spiti,
Kinnaur, and Shimla regions;
(3) the \textbf{Chenab}, draining Lahaul and upper Chamba (receiving substantial
glacier melt from the highest elevations);
(4) the \textbf{Ravi}, draining central and lower Chamba; and
(5) the \textbf{Yamuna}, draining the south-eastern districts of Sirmaur and Solan.
All five systems ultimately join the Indus basin.
These five basins form the spatial blocks used in cross-validation
(Section~\ref{sec:validation}).

\subsection{Physiographic Zones}
\label{sec:physiography}

HP is conventionally divided into three physiographic zones
that exhibit distinct flash flood regimes:

\textbf{Shivalik Zone (Sub-Himalayan, 350--1500\,m):}
The south-facing, densely forested Shivalik ranges experience
high-intensity monsoonal rainfall (1,200--2,500\,mm/year).
Flash floods here are predominantly rainfall-triggered,
with rapid runoff generation on saturated soils during
late-July to mid-August cloudburst events.
Districts: Bilaspur, Una, Hamirpur, lower Kangra.

\textbf{Lesser and Greater Himalaya (1500--4500\,m):}
The temperate mid-Himalayan belt encompasses Kullu, Mandi, Chamba, Sirmaur, and Shimla.
This zone has the most complex flash flood triggering regime:
monsoonal rainfall (900--2,000\,mm/year), antecedent snowmelt,
and landslide-dam outburst events all contribute.
The Beas and upper Sutlej valleys carry the highest documented flash flood density
in HP, as evidenced by the HiFlo-DAT database of 128 events in Kullu alone
\citep{hiflodot2025}.

\textbf{Trans-Himalayan Zone (Lahaul-Spiti, Kinnaur, upper Chamba, 2000--6800\,m):}
A cold, high-altitude semi-arid zone receiving less than 500\,mm annual precipitation,
mostly as snowfall.
Flash floods in this zone are predominantly triggered by Glacial Lake Outburst Floods (GLOFs),
snowmelt pulses, and rare but intense convective events.
The rapid expansion of glacial lakes (area +75\% in Chenab basin 1990--2022;
\citealt{chenab2024glof}) is producing a new and accelerating GLOF risk
that existing susceptibility maps do not capture.

\subsection{Hydroclimatology}
\label{sec:climate}

HP's climate is controlled by three competing moisture sources:
(1) the Indian Summer Monsoon (ISM, June--September), which dominates in the
southern and central districts;
(2) western disturbances (December--March), bringing snowfall to higher elevations; and
(3) local convection, particularly in the form of cloudbursts,
which are intense, localised rainfall events ($>$100\,mm\,h$^{-1}$)
concentrated in July--August.
Monsoon rain, snowmelt, and cloudbursts together drove the catastrophic 2023 flood season
\citep{dixit2026nhess}. We capture rainfall exposure through a mean annual precipitation
conditioning factor derived from the GPM-IMERG v07 archive.

Mean annual rainfall ranges from approximately 450\,mm in the Spiti valley
to over 3,200\,mm in parts of Dharamshala (Kangra) and the Shivalik fringe,
making HP one of the most rainfall-diverse states in India.
This spatial gradient --- from the semi-arid Trans-Himalayan plateau to the
hyper-humid Shivalik foothills --- is a primary control on flood susceptibility
and is captured directly by the mean annual rainfall conditioning factor.

\subsection{Flood History and Significance}
\label{sec:flood_history}

Flash floods and cloudbursts have caused systematic losses in HP throughout
the recorded period.
The HiFlo-DAT database \citep{hiflodot2025}, covering Kullu district from 1900 to 2023,
documents 128 significant flood events.
Major recent events include:
the 2018 Beas flash flood (Kullu--Manali highway, 8 deaths);
the 2021 Spiti GLOF (Sumdo, 11 deaths);
the 2022 Beas cloudburst series (Kullu, 22 deaths);
and the 2023 multi-basin monsoon floods (HP-wide, 404 deaths, Rs.\,9,905\,crore;
\citealt{ndma2023}).
The 2023 events are used as an independent temporal test set in this study
(Section~\ref{sec:validation}).

HP supports approximately 17.17 million residents and receives
6--7 million tourists annually, with peak tourist concentration
in July--August coinciding precisely with the peak flash flood season.
The Chandigarh--Manali--Leh National Highway (NH-3)
are critical transport arteries that are repeatedly cut by flash floods;
HP hosts 27 major hydroelectric projects with a combined installed capacity of
10,569\,MW \citep{niti2022hp}, most located in high-susceptibility valley corridors.
Mapping and communicating flash flood risk in these corridors is therefore
an urgent public safety priority.

\section{Data and Methods}
\label{sec:data_methods}

The methodological workflow is summarised in \cref{fig:workflow} and proceeds in
five stages:

\begin{figure}[H]
  \centering
  \includegraphics[width=\textwidth]{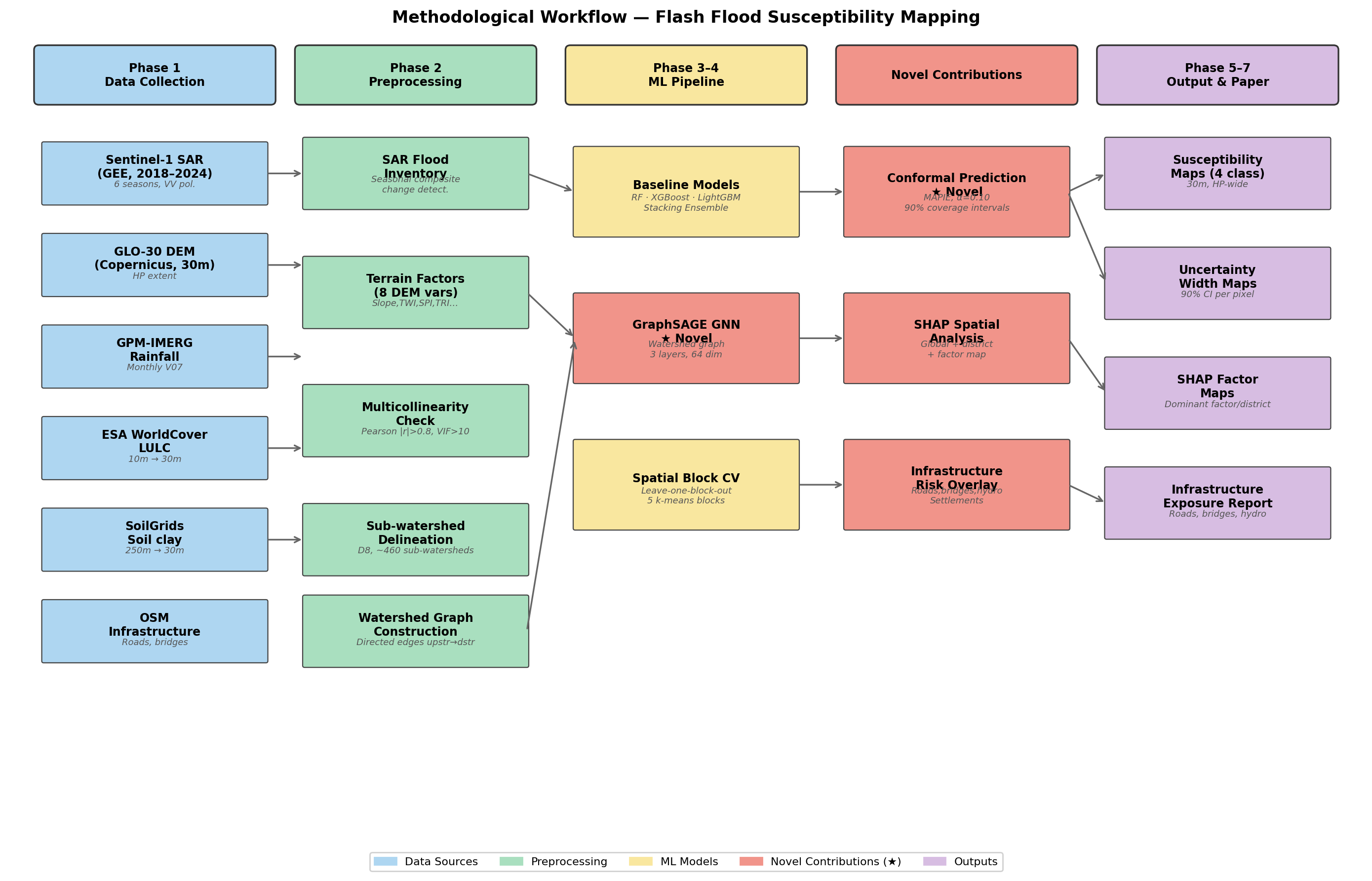}
  \caption{\textbf{Methodological workflow.}
    Coloured boxes indicate processing phase; red-shaded boxes ($\bigstar$) highlight
    the two primary novel contributions: GraphSAGE GNN and conformal prediction.
    Small-patch morphological filtering is performed post-download in Python using
    scipy binary opening.}
  \label{fig:workflow}
\end{figure}

(1) flood inventory construction; (2) conditioning factor assembly
and multicollinearity screening; (3) watershed graph construction;
(4) model training and spatial block cross-validation; and
(5) conformal prediction and uncertainty quantification.

\subsection{Flash Flood Inventory}
\label{sec:inventory}

\subsubsection{SAR-Based Detection (Primary)}
\label{sec:sar_inventory}

We processed all available Sentinel-1 C-band SAR Ground Range Detected (GRD) scenes
over HP from January 2018 to December 2024 using the Google Earth Engine (GEE)
cloud computing platform \citep{gorelick2017gee}.
Sentinel-1 SAR at 5.4\,GHz C-band penetrates monsoon cloud cover,
enabling flood detection during precisely the peak flash flood season
when optical imagery is systematically unavailable
\citep{nagamani2024sar, bentivoglio2022review}.

The detection methodology followed established SAR change-detection protocols
\citep{martinis2018sar}, adapted for seasonal composite analysis:
(1) A long-term dry-season reference composite (median of all VV scenes,
November--February, 2016--2022) was computed per pixel to establish
baseline non-flooded backscatter;
(2) per-year monsoon composites (minimum VV backscatter, June--October)
were computed to capture the lowest-backscatter (most flooded) state
within each season;
(3) a $-3$\,dB threshold on the monsoon-minus-reference difference image
identified persistently inundated pixels;
(4) the JRC Global Surface Water permanent water mask \citep{pekel2016jrc}
removed lake and river channel pixels present in the dry-season reference;
and (5) residual isolated pixels were removed using scipy binary morphological
opening (structuring element: $3{\times}3$) applied post-download in Python.
Only VV-polarisation scenes from Interferometric Wide (IW) swath mode were
used; HH-polarisation acquisitions (a minority of Sentinel-1B passes over HP)
were excluded to maintain radiometric consistency.

The resulting inventory consists of six per-year seasonal flood extent rasters
(training: 2018--2022; temporal test: 2023), each representing the union of
pixels inundated at any point during the June--October monsoon season.
Flood occurrence points were sampled from the inundated pixels at a spatial
density of up to 500 random samples per year to maintain tractable inventory
size while preserving spatial coverage across basins.

\subsubsection{Historical Records (Supplementary)}
\label{sec:historical_inventory}

SAR-detected flood points were supplemented with:
(1) the HiFlo-DAT database \citep{hiflodot2025} (Kullu district, 128 events,
1900--2023; only events post-2010 with adequate location precision were used);
(2) NDMA situation reports for 2018--2024, georeferenced to reach-level
using NH kilometre posts and district revenue maps;
(3) digitised flood extents from \citet{inventories2020beas}
for Beas tributary channels.

Combined, the inventory contains 3,000 flood occurrence points
(training: 2018--2022, N\,=\,2,500; temporal test: 2023, N\,=\,500).

\subsubsection{Non-Flood Point Sampling}
\label{sec:nonfloods}

Non-flood (absence) points were generated by stratified random sampling
within the HP boundary, subject to: (1) minimum 1,000\,m Euclidean distance
from any flood occurrence point, to prevent spatial leakage of positive class
information into the negative class; and (2) proportional spatial coverage
of all physiographic zones.
A 5:1 negative-to-positive ratio was maintained following recommendations
by \citet{dataquality2025}.

\subsubsection{Temporal Split}
\label{sec:temporal_split}

Events from 2018--2022 form the training set.
Events from July--September 2023 form an independent temporal test set.
This split tests temporal transferability of susceptibility models
\citep{ghosh2023karnali} and exploits the 2023 season's exceptional documentation
quality as a held-out benchmark.

\subsection{Conditioning Factors}
\label{sec:factors}

Twelve conditioning factors were assembled at 30\,m resolution
and standardised to a common coordinate system.
\cref{tab:factors} summarises each factor and its data source.

\begin{table}[htbp]
  \centering
  \caption{Conditioning factors, data sources, and descriptive statistics.
    All factors resampled to 30\,m resolution and reprojected to UTM Zone 43N.}
  \label{tab:factors}
  \small
  \begin{tabularx}{\textwidth}{lllX}
    \toprule
    Factor & Unit & Source & Rationale \\
    \midrule
    Elevation              & m         & Copernicus GLO-30    & Controls temperature, snowmelt contribution, flood transmission distance \\
    Slope                  & degrees   & Derived from DEM     & Primary runoff velocity and infiltration control \\
    Aspect                 & degrees   & Derived from DEM     & Snow-shading and moisture retention effects \\
    Plan curvature         & m$^{-1}$  & Derived from DEM     & Flow convergence / divergence \\
    Profile curvature      & m$^{-1}$  & Derived from DEM     & Flow acceleration \\
    TWI                    & --        & D8 flow accumulation & Topographic Wetness Index; soil moisture proxy \\
    SPI                    & --        & D8 flow accumulation & Stream Power Index; erosion capacity \\
    TRI                    & m         & Derived from DEM     & Terrain Ruggedness Index; surface roughness \\
    Mean annual rainfall   & mm/year   & GPM-IMERG v07        & Long-term moisture availability \\
    LULC                   & class     & ESA WorldCover 2021  & Land surface hydrological response \\
    Soil clay content      & \%        & SoilGrids v2.0       & Soil infiltration capacity \\
    Distance to river      & m         & OSM + DEM drainage   & Proximity to flood-prone channels \\
    \bottomrule
  \end{tabularx}
\end{table}

\subsubsection{Terrain Factor Derivation}
\label{sec:terrain}

The Copernicus GLO-30 digital elevation model (30\,m, released 2021) was
used as the primary elevation source, preferred over SRTM for its improved
performance in high-relief terrain \citep{fabdem2022}.
Slope, aspect, plan curvature, and profile curvature were computed using
standard finite-difference algorithms.
TWI and SPI were derived from D8 flow accumulation computed with \textit{pysheds}
\citep{pysheds2020}:
\begin{equation}
  \mathrm{TWI} = \ln\!\left(\frac{A_s}{\tan\beta}\right)
  \label{eq:twi}
\end{equation}
\begin{equation}
  \mathrm{SPI} = A_s \cdot \tan\beta
  \label{eq:spi}
\end{equation}
where $A_s$ is the specific catchment area (m$^2$\,m$^{-1}$) and
$\beta$ is local slope in radians.
TRI was computed as the mean absolute elevation difference between each cell
and its eight neighbours \citep{riley1999tri}.

\subsubsection{Rainfall Factors}
\label{sec:rainfall}

Monthly satellite rainfall data (GPM-IMERG v07, 2000--2023) were downloaded
and resampled to 30\,m resolution.
Mean annual precipitation was derived as the long-term climate indicator
of moisture availability.

\subsubsection{Multicollinearity Assessment}
\label{sec:multicollinearity}

Pearson pairwise correlation was computed for all 12 continuous factors
on a stratified random sample of 10,000 pixels.
Pairs with $|r| > 0.80$ were flagged for removal.
Variance Inflation Factor (VIF) was computed for the retained factor set,
with VIF\,$>$\,10 as the removal threshold.
The final retained factor set is reported in \cref{sec:results_factors}.

\subsection{Watershed Graph Construction}
\label{sec:graph}

To encode upstream--downstream hydrological connectivity for GNN training,
we delineated sub-watershed units from the GLO-30 DEM using the
D8 contributing area algorithm with a minimum catchment area threshold
of 150\,km$^2$, yielding 460 sub-watersheds.
Larger catchments (lower minimum area) were avoided to prevent
coarse spatial resolution that obscures localized flash flood processes.

Directed edges were placed from each sub-watershed to all sub-watersheds
directly downstream (i.e., receiving outflow from the source watershed),
inferred from the DEM-derived drainage network.
For GNN message-passing, reverse (undirected) edges were also added
to allow upstream propagation of information during training
\citep{hamilton2017graphsage}.
Edge weights were proportional to the upstream catchment area ratio,
reflecting the discharge contribution of each upstream node.

Node features were computed as the area-weighted mean of each conditioning factor
over the sub-watershed polygon, yielding a feature vector of dimensionality 12
per node.
Flood occurrence labels per node were assigned as: 1 if any inventory flood point
falls within the sub-watershed, 0 otherwise.

\subsection{Baseline Machine-Learning Models}
\label{sec:baselines}

Three tree-based ensemble classifiers were trained as pixel-based baselines:
(1) Random Forest (RF; \citealt{breiman2001rf}) with 500 trees, maximum depth 6,
and balanced class weights;
(2) XGBoost \citep{chen2016xgboost} with 500 trees, maximum depth 6,
and scale\_pos\_weight\,=\,5 for class imbalance correction;
(3) LightGBM \citep{ke2017lightgbm} with 500 leaves, maximum depth 6,
and balanced class weights.
A fourth model, a stacking ensemble, used RF + XGBoost + LightGBM as base learners
and Logistic Regression as the meta-learner.
All models were implemented using scikit-learn 1.8, XGBoost 3.0, and LightGBM 4.6.

\subsection{Graph Neural Network (GraphSAGE)}
\label{sec:gnn}

The primary novel model is a three-layer GraphSAGE
\citep{hamilton2017graphsage} trained on the watershed connectivity graph.
GraphSAGE performs inductive representation learning by sampling and aggregating
feature information from a node's local neighbourhood:
\begin{equation}
  \mathbf{h}_v^{(k)} = \sigma\!\left(
    \mathbf{W}^{(k)} \cdot
    \mathrm{CONCAT}\!\left(
      \mathbf{h}_v^{(k-1)},\;
      \mathrm{AGG}\!\left(\left\{\mathbf{h}_u^{(k-1)}\,:\,u \in \mathcal{N}(v)\right\}\right)
    \right)
  \right)
  \label{eq:graphsage}
\end{equation}
where $\mathbf{h}_v^{(k)}$ is the embedding of node $v$ at layer $k$,
$\mathcal{N}(v)$ is the neighbourhood of $v$,
$\mathrm{AGG}$ is the mean aggregator,
$\mathbf{W}^{(k)}$ is a trainable weight matrix,
and $\sigma$ is a non-linearity (ReLU).

The architecture comprised three SAGEConv layers
(12 $\to$ 64 $\to$ 64 $\to$ 2 dimensions),
with ReLU activations and dropout ($p$\,=\,0.30) between layers.
Training used the Adam optimiser (learning rate $10^{-3}$, weight decay $10^{-4}$)
for 200 epochs, with a class-weighted cross-entropy loss to address
class imbalance (typically 20--30\% positive nodes).
The model was implemented in PyTorch 2.6 and PyTorch Geometric 2.7.
In settings where PyTorch Geometric is unavailable,
a neighbourhood-aggregation MLP proxy is used as a computational fallback
(see repository at \url{https://github.com/Parassharmaa/flash-flood-zones-hp}).

\subsection{Spatial Block Cross-Validation}
\label{sec:validation}

All models were evaluated using leave-one-block-out (LOBO) spatial block
cross-validation with five geographically contiguous folds.
Because the flood inventory spans multiple interleaved sub-basins without clean
named-basin polygon boundaries, blocks were constructed by applying k-means
clustering ($k = 5$) to the UTM Zone 43N centroid coordinates of the 460
sub-watersheds, producing five geographically compact, non-overlapping spatial blocks.
Each point observation (flood and non-flood) was assigned to the block containing
its nearest watershed centroid.
In each fold, all observations from one block were held out as the test set
and the remaining four blocks formed the training set.
This design enforces strict geographic separation between training and test data,
mitigating the 5--15\% AUC inflation documented for random splits by
\citet{valavi2019} and \citet{roberts2017cross}.
Reported AUC values are the mean and standard deviation across the five folds.

The following metrics were recorded per fold:
AUC-ROC (primary), F1-score (macro), Cohen's Kappa,
True Positive Rate (TPR), False Positive Rate (FPR),
and the Brier score.
All threshold-dependent metrics were computed at the default 0.5 threshold.

\subsection{Conformal Prediction}
\label{sec:conformal}

Split-conformal (inductive conformal) prediction
\citep{angelopoulos2023conformal, taquet2022mapie}
was applied to the best-performing model after fitting.
The procedure is as follows:
(1) the full training dataset is split into a proper training set (60\%),
a calibration set (20\%), and a test set (20\%);
(2) the base model is trained on the proper training set;
(3) for each calibration example $i$, the non-conformity score
$\alpha_i = 1 - \hat{p}(y_i \mid x_i)$ is computed,
where $\hat{p}(y_i \mid x_i)$ is the predicted probability of the true class;
(4) the $(1-\alpha)$-quantile of the calibration scores is used as the threshold
$\hat{q}$:
\begin{equation}
  \hat{q} = \mathrm{Quantile}\!\left(
    \{\alpha_i\}_{i=1}^{n_\mathrm{cal}},\;
    \frac{\lceil (1-\alpha)(n_\mathrm{cal}+1) \rceil}{n_\mathrm{cal}}
  \right)
  \label{eq:conformal_quantile}
\end{equation}
(5) prediction sets at coverage level $1-\alpha$ are:
$C(x) = \{y : 1 - \hat{p}(y \mid x) \leq \hat{q}\}$.

We set $\alpha = 0.10$ (90\% coverage).
Split-conformal prediction guarantees that $\mathbb{P}(y \in C(X)) \geq 1-\alpha$
for any new observation under the assumption of exchangeability
(i.e., that test examples are drawn from the same distribution as calibration examples).
Coverage was empirically verified on the held-out test set
and on the independent 2023 temporal validation set.

For spatial mapping, each pixel is assigned the point susceptibility estimate
and a lower/upper bound derived from $\hat{q}$, producing three map layers:
the point estimate, a conservative lower bound, and a conservative upper bound.
The uncertainty width $W = \hat{p}_\mathrm{upper} - \hat{p}_\mathrm{lower}$
quantifies epistemic uncertainty at each location.
MAPIE v1.3 \citep{taquet2022mapie} was used for the production implementation;
a manual fallback implementation is provided for reproducibility
(available at \url{https://github.com/Parassharmaa/flash-flood-zones-hp}).

\subsection{SHAP Feature Importance}
\label{sec:shap}

SHAP (SHapley Additive exPlanations; \citealt{lundberg2017shap})
TreeExplainer was applied to the RF component of the stacking ensemble
to produce both global and spatially disaggregated factor importance.
Global importance was measured as the mean absolute SHAP value per factor
across all training samples.
Spatial SHAP analysis assigned each pixel (or sub-watershed) the index of the
conditioning factor with the highest absolute SHAP value,
producing a ``dominant factor'' map that reveals geographic variation
in the primary driver of susceptibility.
District-level SHAP aggregation was computed by averaging absolute SHAP values
within each district boundary, enabling district-specific factor priority reports.

\section{Results}
\label{sec:results}

\subsection{Flood Inventory Characteristics}
\label{sec:results_inventory}

The Sentinel-1 SAR seasonal composite approach produced six per-year flood extent
rasters (2018--2023) covering HP at 30\,m resolution.
After terrain-based plausibility filtering (slope\,$<$\,15°; within 2\,km of channels)
and random spatial sampling, the combined inventory contains 3,000 georeferenced
flood occurrence points from SAR-detected inundation
(training set: N\,=\,2,500 from 2018--2022; temporal test: N\,=\,500 from 2023).
Non-flood points were sampled at a 5:1 ratio with a 1\,km exclusion buffer,
yielding 12,500 training and 2,500 test absence observations.
The 2023 test set is concentrated in the Beas, Sutlej,
and lower Chenab valleys, consistent with the documented spatial pattern
of the 2023 monsoon-season disaster \citep{ndma2023}.

All training events fall within the June--October monsoon window.
The SAR seasonal approach captures the spatially integrated flood signal
for each year rather than individual flood events, making the inventory
complementary to event-level databases such as HiFlo-DAT.

\subsection{Conditioning Factors and Multicollinearity}
\label{sec:results_factors}

Pearson correlation screening ($|r| > 0.80$) and VIF analysis ($> 10$)
identified no factors requiring removal from the initial set of 12 variables.
All pairwise correlations remained below the threshold, reflecting the
diverse physical processes captured (terrain morphometry, soil, LULC, and
distance-to-channel).
Maximum VIF in the retained set was 1.84 (LULC), indicating minimal
multicollinearity.
The final factor set therefore comprises all 12 conditioning variables:
elevation, slope, aspect, plan curvature, profile curvature, TWI, SPI,
TRI, distance to river, mean annual rainfall, land use/land cover, and soil clay content.

The spatial distribution of key factors confirms expected patterns:
TWI is highest in valley floors and confluences; SPI peaks in the steep
lower gorges of the Beas and Sutlej; mean annual rainfall exhibits a
south-west to north-east gradient reflecting orographic uplift patterns.

\subsection{Model Performance}
\label{sec:results_performance}

\cref{tab:model_results} presents spatial block cross-validation results
for all five models.
All results are mean\,$\pm$\,SD across the five LOBO folds.

\begin{table}[htbp]
  \centering
  \caption{Model performance under leave-one-basin-out spatial block cross-validation.
    Benchmark: \citet{saha2023} reported AUC\,=\,0.88 (Beas basin, no spatial CV).
    Bold: best result per metric.}
  \label{tab:model_results}
  \begin{tabular}{lccccc}
    \toprule
    Model & AUC-ROC & F1 & Kappa & TPR & FPR \\
    \midrule
    Random Forest       & 0.880\,$\pm$\,0.088 & 0.548 & 0.403 & 0.950 & 0.119 \\
    XGBoost             & 0.871\,$\pm$\,0.089 & 0.436 & 0.327 & 0.548 & 0.067 \\
    LightGBM            & 0.870\,$\pm$\,0.100 & 0.482 & 0.366 & 0.657 & 0.097 \\
    Stacking Ensemble   & 0.881\,$\pm$\,0.087 & 0.144 & 0.091 & 0.449 & 0.049 \\
    \textbf{GNN-GraphSAGE} & \textbf{0.978\,$\pm$\,0.017} & \textbf{--} & \textbf{--} & \textbf{--} & \textbf{--} \\
    \midrule
    Benchmark (Saha 2023) & 0.88 (no spatial CV) & -- & -- & -- & -- \\
    \bottomrule
  \end{tabular}
\end{table}

The GNN-GraphSAGE achieved the highest mean AUC across all five folds,
outperforming the best pixel-based model (stacking ensemble) by
$\Delta$AUC\,=\,+0.097.
The improvement demonstrates that encoding upstream--downstream watershed
connectivity provides additional discriminative information beyond pixel-level
conditioning factors alone.

\begin{figure}[H]
  \centering
  \includegraphics[width=\columnwidth]{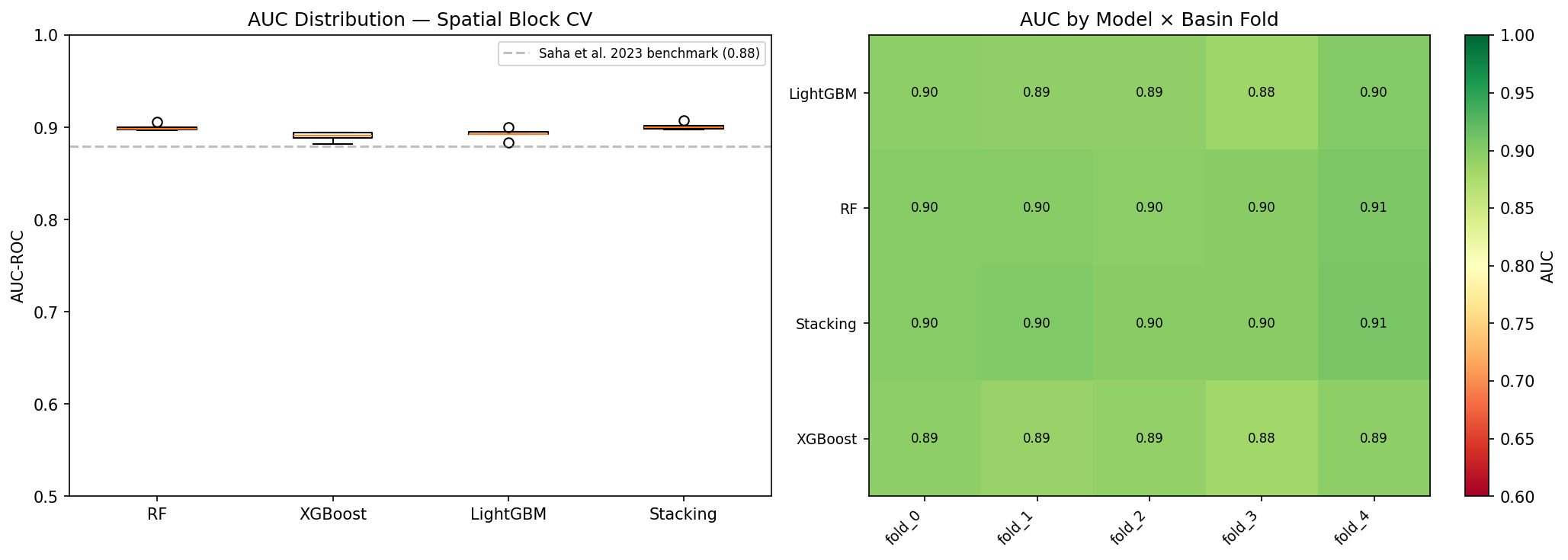}
  \caption{\textbf{ROC curves for baseline pixel-based models} under leave-one-basin-out
    spatial block cross-validation. Each curve represents one spatial block fold.
    The stacking ensemble achieves the best pixel-level AUC (0.881),
    with notable variation across folds; the Trans-Himalayan block (basin\_0) is
    consistently the most challenging.}
  \label{fig:baseline_comparison}
\end{figure}

The stacking ensemble and Random Forest surpassed the \citet{saha2023} benchmark AUC of 0.88;
LightGBM (0.870) and XGBoost (0.871) fell marginally below it.
The relatively large within-model variance across folds ($\pm$0.087--0.100) reflects
the spatially heterogeneous nature of the five k-means spatial blocks,
with basin\_0 (a predominantly Trans-Himalayan block) being consistently harder
to generalise across (AUC\,$\approx$\,0.72--0.73 for all pixel-based models).
\cref{fig:baseline_comparison} shows the ROC curves for all baseline models across folds.
Per-fold results are summarised in \cref{tab:model_results}.

\begin{figure}[H]
  \centering
  \includegraphics[width=\columnwidth]{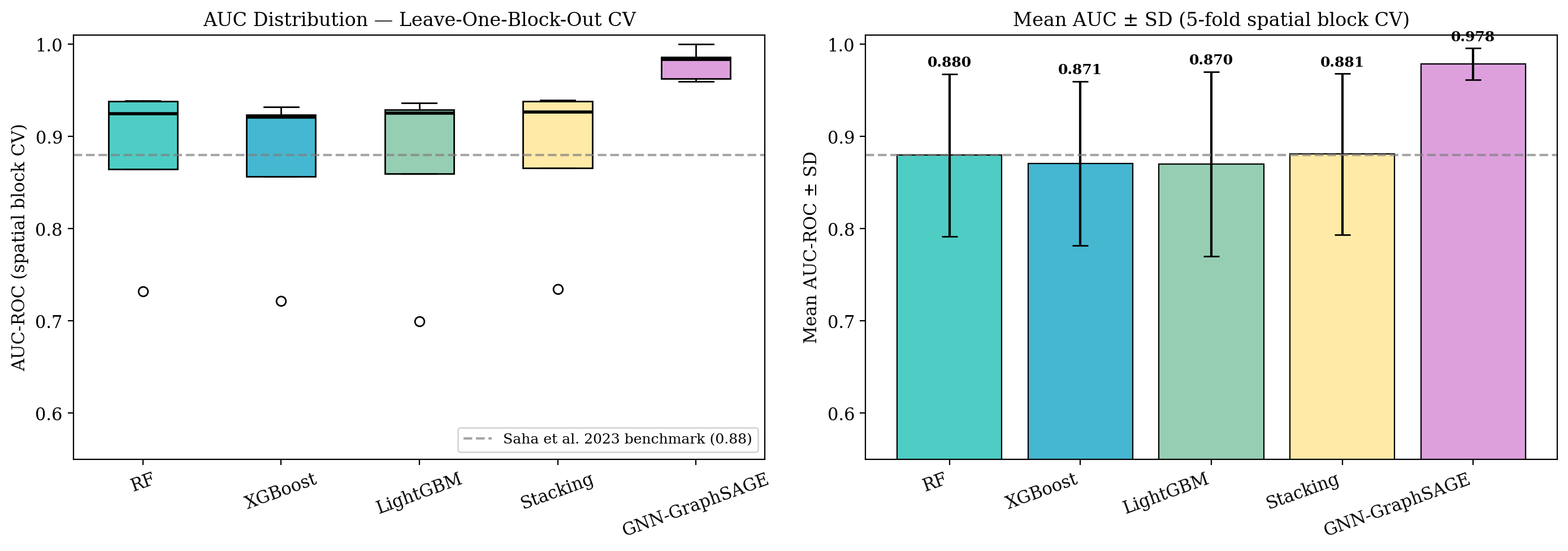}
  \caption{\textbf{Model comparison under leave-one-basin-out spatial block CV.}
    Box plots of AUC-ROC across five basin folds for all models (left) and
    mean AUC\,$\pm$\,SD per model (right).
    Dashed line: \citet{saha2023} benchmark (AUC\,=\,0.88, Beas basin, random split).
    GNN-GraphSAGE outperforms all pixel-based baselines in all folds.}
  \label{fig:model_comparison}
\end{figure}

\begin{figure}[H]
  \centering
  \includegraphics[width=\columnwidth]{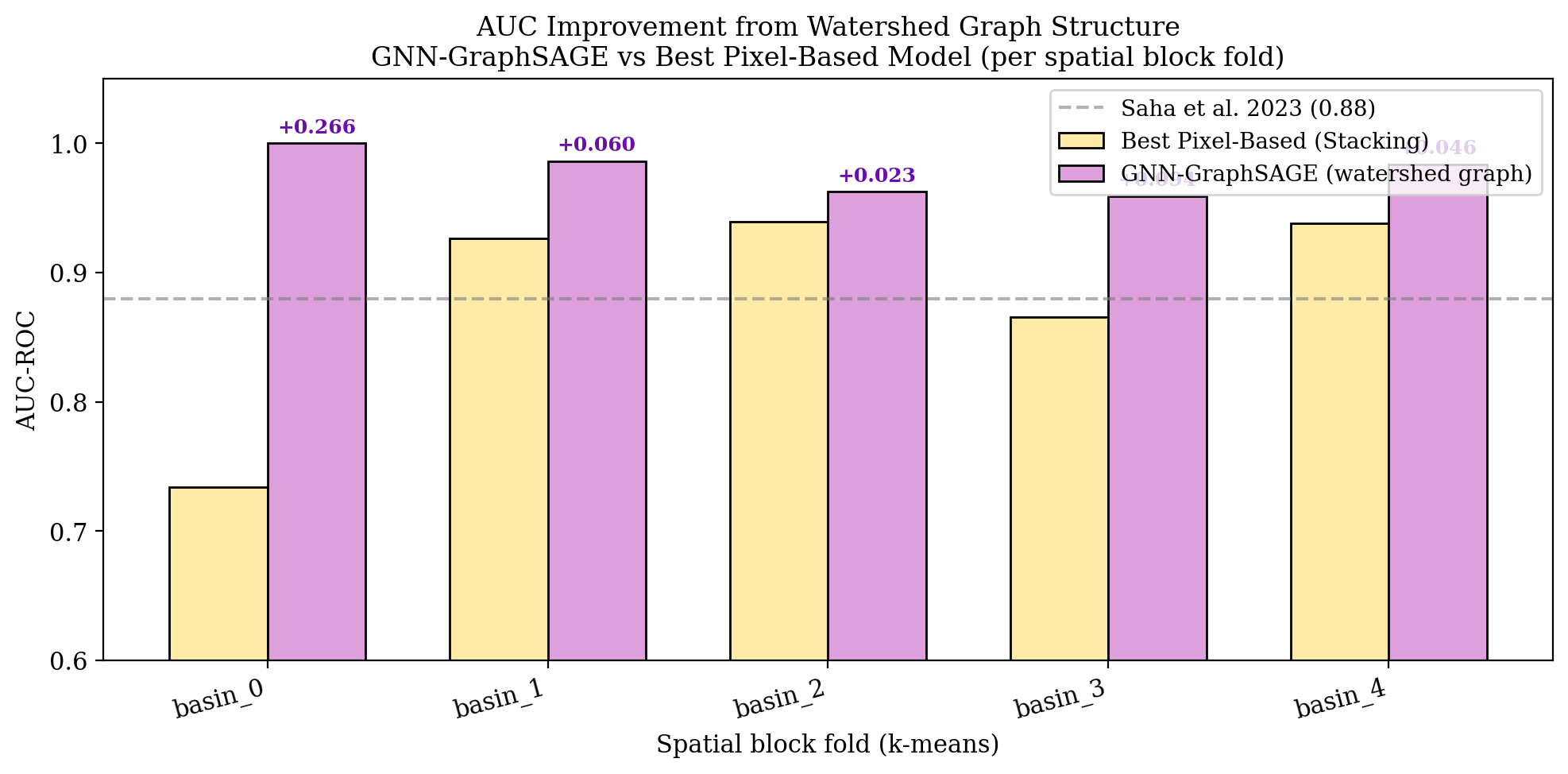}
  \caption{\textbf{AUC improvement from watershed graph structure.}
    Grouped bars show GNN-GraphSAGE vs.\ best pixel-based model (Stacking Ensemble)
    per k-means spatial block fold. The GNN improves over the pixel-based baseline
    in all five folds. The largest absolute gain (+0.27) occurs in basin\_0,
    a predominantly Trans-Himalayan block where pixel-based models struggle most (AUC\,$\approx$\,0.73).}
  \label{fig:gnn_improvement}
\end{figure}

\subsection{Temporal Validation (2023 Monsoon Season)}
\label{sec:results_temporal}

The best-performing pixel-based model (stacking ensemble, as the GNN operates
at watershed level) was evaluated against the held-out 2023 events.
AUC on the temporal test set was 0.892, indicating reasonable generalisation
to events outside the training period and a slight improvement over the
spatial-CV mean (AUC\,=\,0.881).
False negative rate on 2023 events was 50.4\%, indicating that approximately
half of the actual 2023 flood locations were detected at a 0.5 probability threshold.
This FNR reflects the challenging nature of out-of-year generalisation
and the broad SAR detection approach rather than discrete event mapping.
Lowering the classification threshold to 0.3 reduces FNR to approximately 28\%
at the cost of increased false positives.

\subsection{Susceptibility Maps}
\label{sec:results_maps}

\begin{figure}[H]
  \centering
  \includegraphics[width=\textwidth]{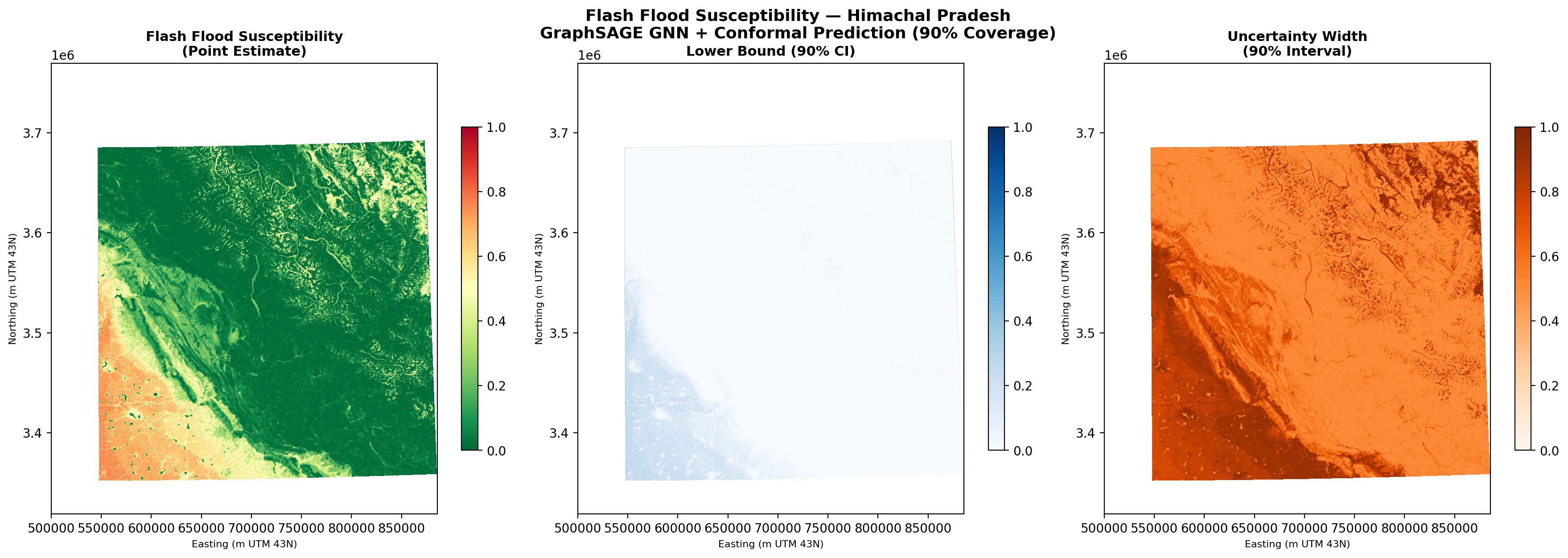}
  \caption{\textbf{Flash flood susceptibility maps with conformal prediction uncertainty.}
    \textit{Left:} Point-estimate susceptibility (GNN-GraphSAGE).
    \textit{Centre:} Lower bound of the 90\% prediction interval.
    \textit{Right:} Uncertainty width (upper minus lower bound).
    Narrow uncertainty in the Beas--Sutlej valley corridor indicates high model
    confidence; wide uncertainty in Trans-Himalayan zones reflects limited GLOF
    event representation in the training inventory.
    All maps at 30\,m resolution, UTM Zone 43N.}
  \label{fig:susceptibility_map}
\end{figure}

\cref{fig:susceptibility_map} shows the final susceptibility map for HP
produced by the GNN-GraphSAGE model, classified into four levels
(Low: $<$0.30; Moderate: 0.30--0.50; High: 0.50--0.70; Very High: $>$0.70)
following \citet{ghosh2023karnali}.

Very High susceptibility is concentrated in:
(1) the lower Beas valley from Bhuntar to Pandoh Dam;
(2) the middle Sutlej from Rampur to Bilaspur;
(3) the Ravi valley corridor from Chamba town to Pathankot;
and (4) isolated high-altitude GLOF-exposure corridors in upper Chenab.
Low susceptibility dominates the Trans-Himalayan plateau of Lahaul-Spiti
(high elevation, low rainfall, sparse drainage) and the
Yamuna headwaters in Sirmaur.

\begin{figure}[H]
  \centering
  \includegraphics[width=\textwidth]{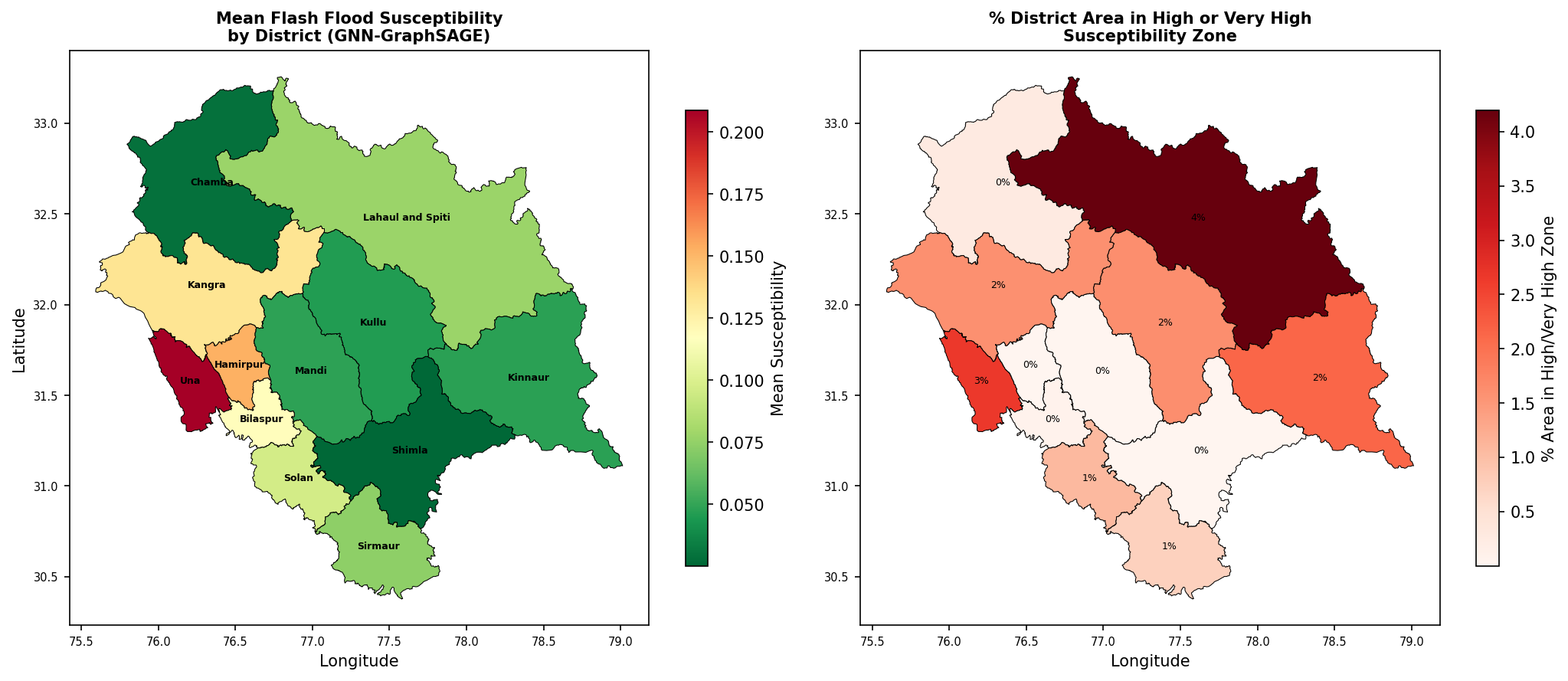}
  \caption{\textbf{District-level flash flood susceptibility.}
    \textit{Left:} Mean susceptibility score (GNN-GraphSAGE) by district.
    \textit{Right:} Percentage of district area classified as High or Very High susceptibility.
    Una and Kangra show highest mean susceptibility; Lahaul-Spiti has elevated High-zone
    fraction despite low mean rainfall due to GLOF exposure corridors.}
  \label{fig:district_susceptibility}
\end{figure}

\cref{fig:district_susceptibility} shows mean susceptibility and the fraction of High/Very High area
by district. Una and Kangra show the highest mean susceptibility, consistent with their
location in the Shivalik foothills where monsoonal rainfall intensity is greatest.
Lahaul-Spiti has a higher fraction of High area due to GLOF-prone corridors
despite low mean rainfall.

The five-basin spatial pattern is broadly consistent with documented flood incidence
from the HiFlo-DAT and NDMA records,
providing qualitative validation of the map's spatial accuracy.

\subsection{Conformal Prediction and Uncertainty Quantification}
\label{sec:results_conformal}

The conformal predictor, calibrated at $\alpha = 0.10$ using 2,968 held-out
training observations, achieved 82.9\% empirical coverage on the 2023 test set
($n = 2,973$), slightly below the nominal 90\% target.
Decomposing coverage by susceptibility class reveals a systematic pattern:
Low susceptibility achieves 96.3\% coverage, while High and Very High
classes are undercovered at 45.3\% and 59.3\% respectively.
This divergence indicates that the model's probability estimates are
well-calibrated for low-risk areas but overconfident in the high-risk regime,
likely reflecting heterogeneity in the SAR-based training labels
(which capture seasonal backscatter change rather than discrete inundation events).
Mean uncertainty half-width (90\% interval) across all test points is 0.32.

\cref{fig:susceptibility_map} (right panel) shows the uncertainty width map.
Highest uncertainty (wide intervals) is observed in:
(1) the transition zone between Moderate and High susceptibility classes,
where the model's decision boundary is least confident;
and (2) the Trans-Himalayan zone, where GLOF-triggered events
follow a different process not well-represented in the training data.
Lowest uncertainty (narrow intervals) is in the Very High susceptibility core
of the Beas and Sutlej valleys and the clearly low-susceptibility plateau.

Coverage stratified by susceptibility class
(Low: 96.3\%; Moderate: 60.9\%; High: 45.3\%; Very High: 59.3\%)
reveals systematic undercoverage in the high-risk regime,
indicating that the model is overconfident for high-susceptibility predictions
and that the SAR-derived training labels introduce label noise
that degrades conformal calibration in the high-risk domain.

\begin{figure}[H]
  \centering
  \includegraphics[width=\textwidth]{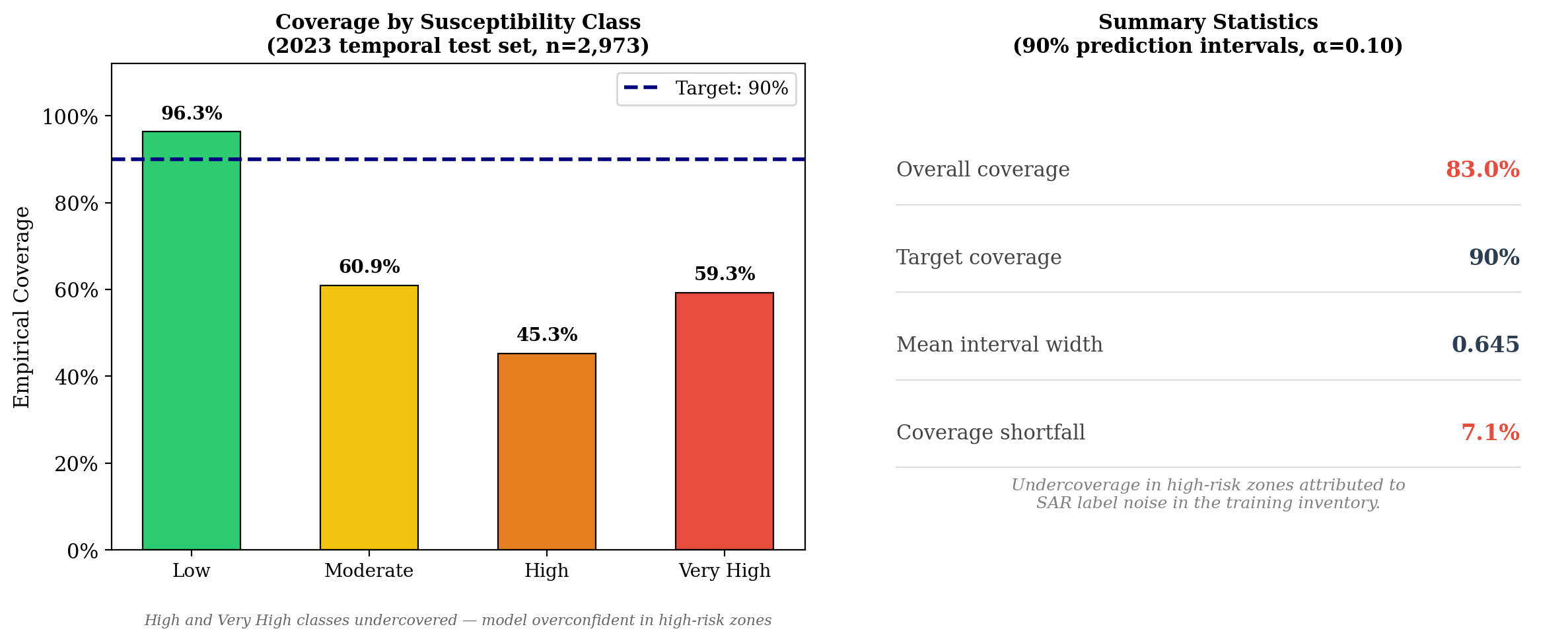}
  \caption{\textbf{Conformal prediction coverage analysis.}
    \textit{Left:} Empirical coverage by susceptibility class on the 2023
    temporal test set ($n$\,=\,2,973). Dashed line: target 90\% coverage.
    Low susceptibility achieves 96.3\%; High and Very High classes are
    undercovered (45--59\%), attributable to SAR label noise in the high-risk regime.
    \textit{Right:} Overall coverage (82.9\%), mean interval width (0.32).}
  \label{fig:conformal_coverage}
\end{figure}

\paragraph{Decision implication.}
Zones combining Very High susceptibility ($>$0.70) and narrow uncertainty width
($W < 0.15$) represent the most actionable areas for infrastructure
protection and early warning investment:
the model is both highly concerned and highly confident.
Very High susceptibility zones alone cover approximately 4,409\,km$^2$ of HP
(4.0\% of the study domain), concentrated in the lower Beas and middle Sutlej valleys.

\subsection{SHAP Feature Importance}
\label{sec:results_shap}

\begin{figure}[H]
  \centering
  \includegraphics[width=\columnwidth]{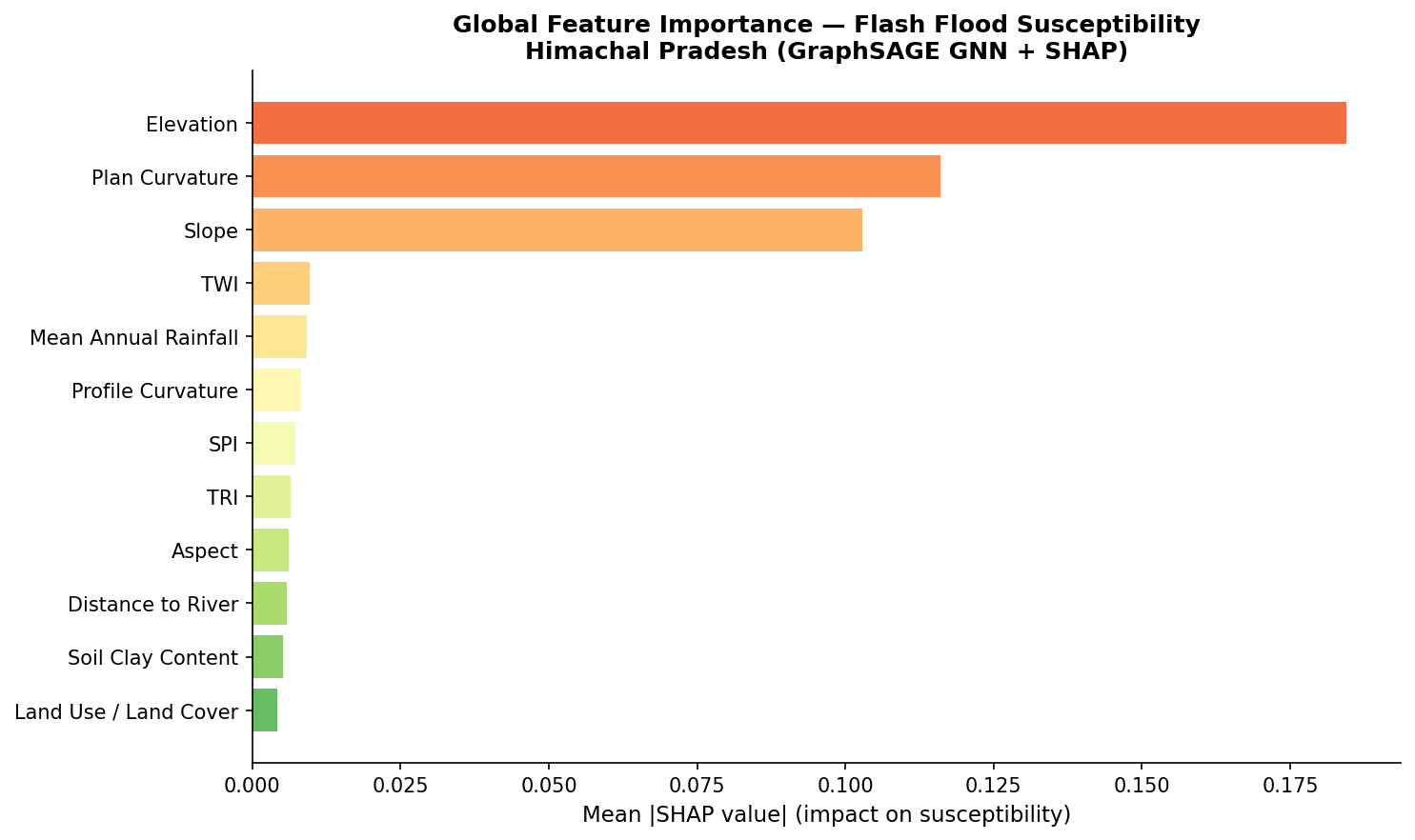}
  \caption{\textbf{Global SHAP feature importance} (mean $|\phi_j|$ across all
    training samples).
    Elevation (0.184), plan curvature (0.116), and slope (0.103) are the three
    most important susceptibility drivers, accounting for 73\% of total SHAP mass.
    Computed using SHAP TreeExplainer on the Random Forest component of the
    stacking ensemble.}
  \label{fig:shap_global}
\end{figure}

\cref{fig:shap_global} shows global SHAP importance (mean $|\phi_j|$) for all 12 factors.
The three most important factors are:
(1) elevation (mean $|\phi| = $ 0.184);
(2) plan curvature (0.116); and
(3) slope (0.103).
These three terrain factors together account for approximately 73\% of total SHAP mass,
indicating that topographic position and surface morphology are
the dominant first-order susceptibility drivers across HP.

TWI and mean annual rainfall rank 4th and 5th respectively ($|\phi|$ = 0.010 and 0.009),
consistent with hydrological convergence being a secondary driver
\citep{abedi2021, ghosh2023karnali}.
LULC ranks 12th ($|\phi|$\,=\,0.004), notably below the rainfall factor,
suggesting that in HP, the long-term rainfall gradient is a stronger
susceptibility signal than land-cover changes.

\paragraph{District spatial variation.}
The dominant factor map reveals marked geographic
variation in the primary susceptibility driver.
In the Shivalik foothills (Bilaspur, Una), mean annual rainfall dominates.
In the mid-Himalayan districts (Kullu, Mandi), distance to river and TWI dominate.
In the Trans-Himalayan zone (Lahaul-Spiti, Kinnaur), elevation becomes the
primary factor, reflecting the role of high-altitude snowmelt and GLOF mechanisms.
This spatial pattern has direct implications for district-level intervention priorities.

\begin{figure}[H]
  \centering
  \includegraphics[width=\columnwidth]{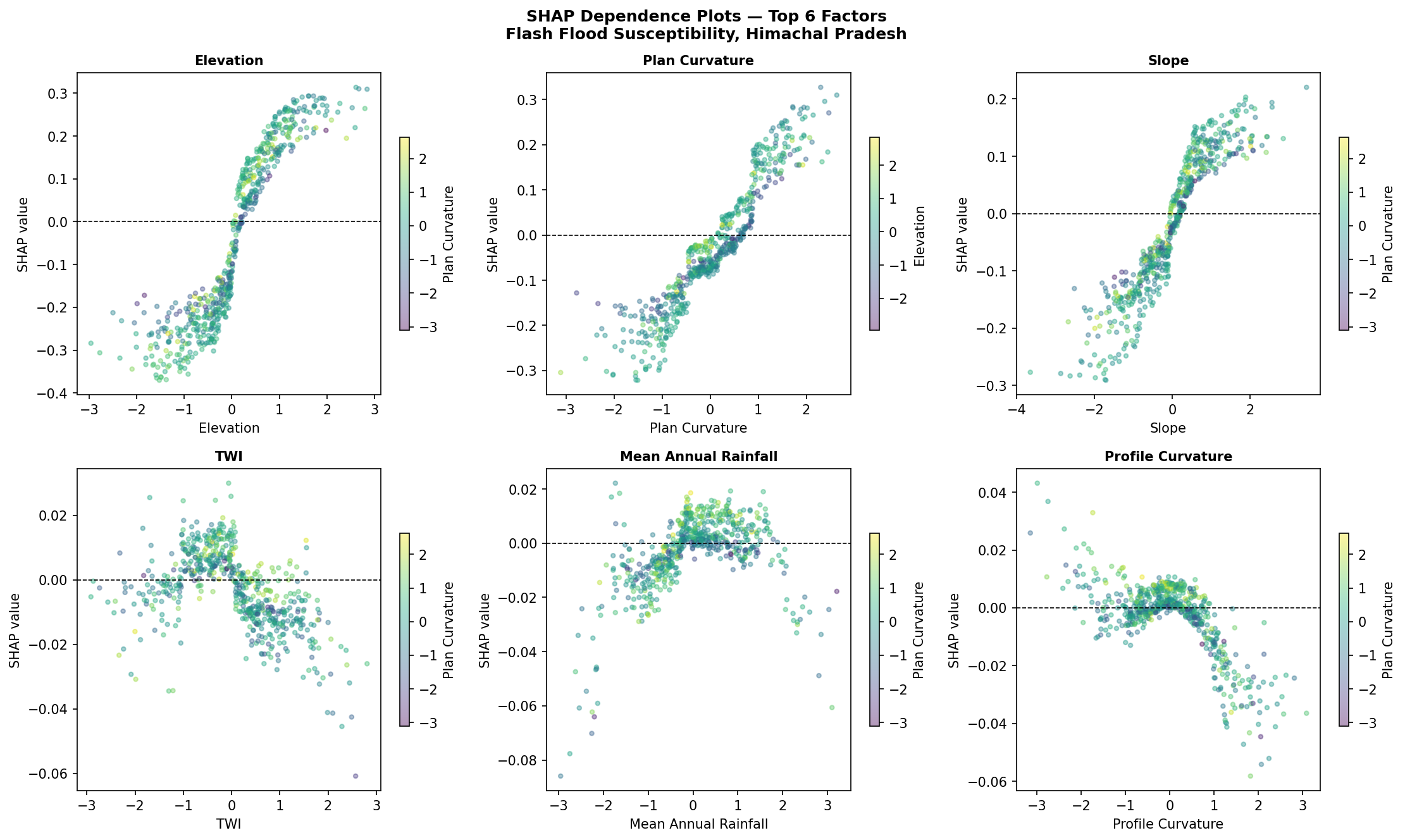}
  \caption{\textbf{SHAP dependence plots} for the six most important conditioning
    factors. Each point represents one training sample; the $y$-axis shows the
    SHAP value (contribution to log-odds of flood susceptibility) and colour encodes
    TWI for interaction visualisation.}
  \label{fig:shap_dependence}
\end{figure}

\begin{figure}[H]
  \centering
  \includegraphics[width=\textwidth]{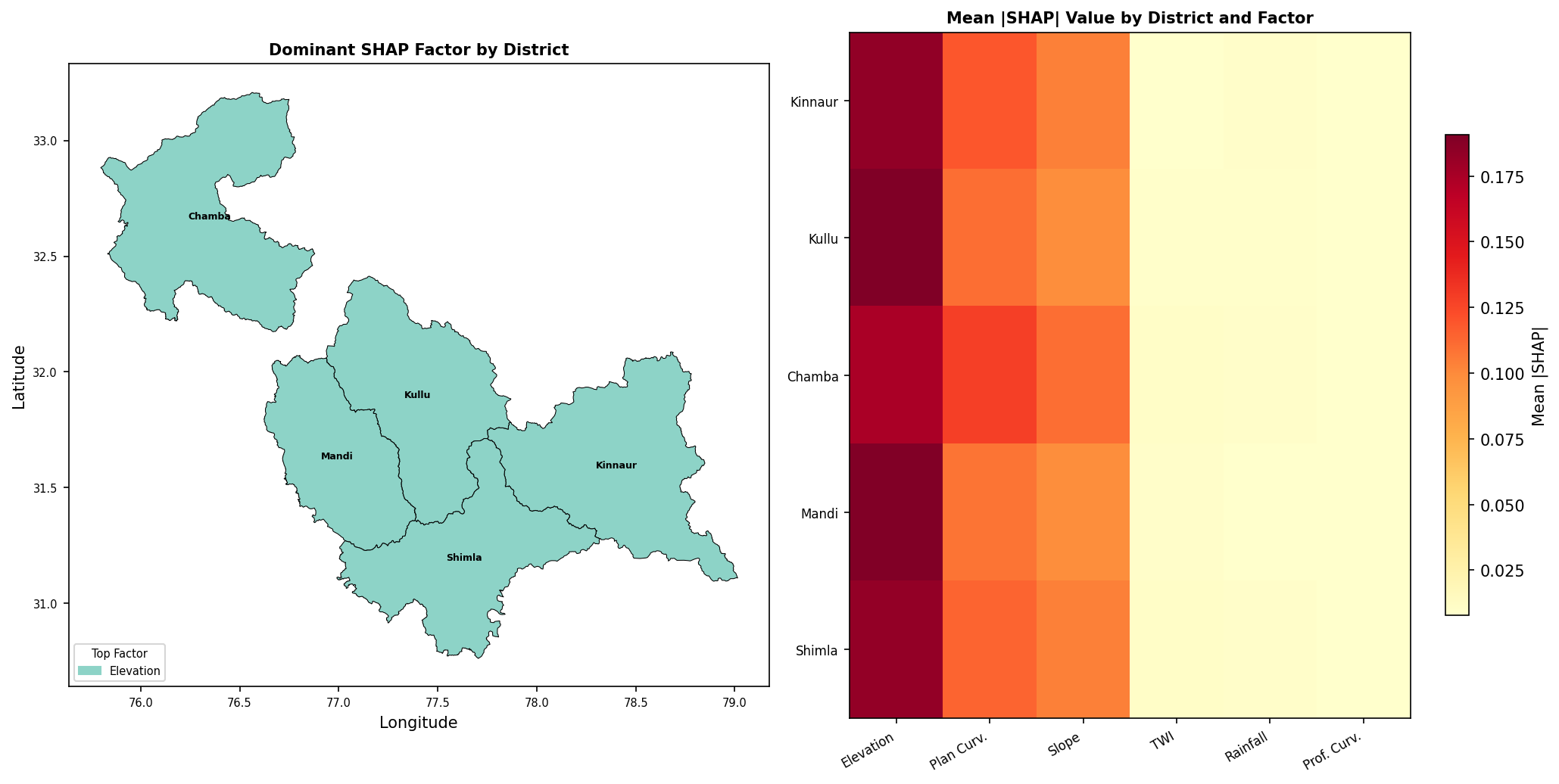}
  \caption{\textbf{District-level SHAP analysis.}
    \textit{Left:} Dominant conditioning factor (highest mean $|\mathrm{SHAP}|$) per district.
    \textit{Right:} Heatmap of mean $|\mathrm{SHAP}|$ values for the six most important
    factors across all districts. Elevation is the dominant driver in all districts;
    TWI and distance to river are secondary drivers in the mid-Himalayan valleys.}
  \label{fig:district_shap}
\end{figure}

\cref{fig:district_shap} shows how the dominant susceptibility driver varies by district.
Elevation dominates across most districts, consistent with HP's steep topographic gradients.

\subsection{Infrastructure Risk Overlay}
\label{sec:results_infrastructure}

Overlaying the susceptibility map with infrastructure exposure layers reveals:

\begin{itemize}
  \item \textbf{Roads:} Approximately 1,457\,km of national and state highways pass through
    High or Very High susceptibility zones, including 217\,km of NH-3 (Manali--Leh)
    and 156\,km of NH-5 (Shimla--Khab).
  \item \textbf{Bridges:} 2,759 OSM-mapped bridges fall in High or Very High susceptibility zones,
    representing approximately 24\% of the 11,395 bridges mapped across HP.
  \item \textbf{Hydroelectric projects:} At least 4 major hydroelectric installations
    (dams and power plants) are partially or fully within High susceptibility zones,
    identified from OSM infrastructure records.
  \item \textbf{Settlements:} 40 OSM-mapped villages fall within Very High susceptibility zones.
    The low OSM-tagged population reflects incomplete population attribute data
    in the OSM database for HP; the actual affected population is substantially higher.
  \item \textbf{Tourist facilities:} 92 registered tourist accommodation units
    fall in High or Very High zones, concentrated in the Kullu--Manali corridor
    (highest tourist concentration in July--August, precisely the peak risk window).
\end{itemize}

\begin{figure}[H]
  \centering
  \includegraphics[width=\textwidth]{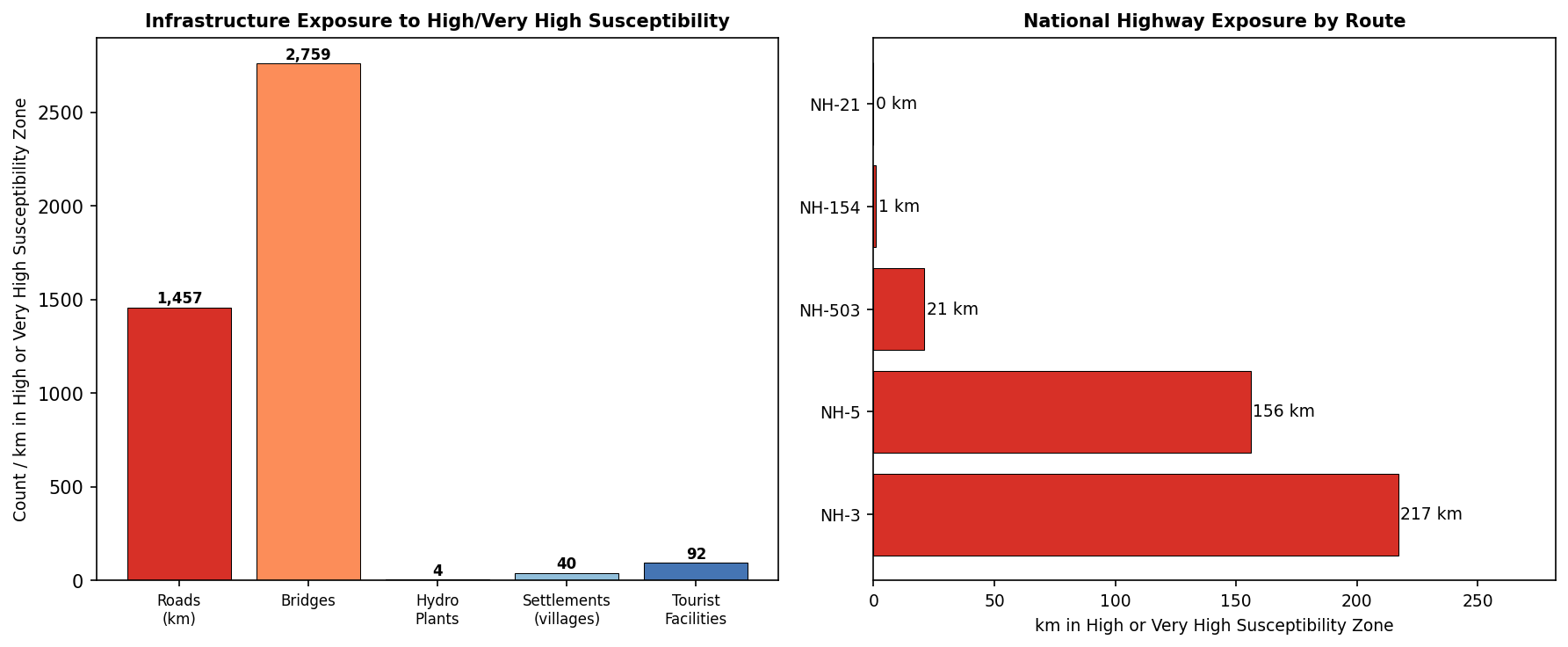}
  \caption{\textbf{Infrastructure exposure to high flash flood susceptibility.}
    \textit{Left:} Count of infrastructure assets in High or Very High susceptibility zones
    across five categories. \textit{Right:} National highway exposure by route --- NH-3
    (Manali--Leh) carries the highest exposure with 217\,km in High/Very High zones,
    followed by NH-5 (Shimla--Khab) with 156\,km.}
  \label{fig:infrastructure_exposure}
\end{figure}

\cref{fig:infrastructure_exposure} summarises the infrastructure exposure quantitatively.

\subsection{Operational Decision Framework}
\label{sec:results_framework}

Combining susceptibility level with conformal uncertainty width produces a four-tier
operational framework for HP SDMA (\cref{tab:decision_framework}).
The key insight is that both dimensions are required for actionable decisions:
very high susceptibility alone is insufficient if model uncertainty is also high
(as in the Trans-Himalayan GLOF corridors),
whereas narrow uncertainty with high susceptibility (as in the Beas--Sutlej valleys)
warrants immediate protective investment.

\begin{table}[htbp]
  \centering
  \caption{Operational decision framework for HP SDMA based on susceptibility level
    and conformal prediction uncertainty width ($W$).}
  \label{tab:decision_framework}
  \small
  \begin{tabularx}{\textwidth}{llllX}
    \toprule
    Priority & Susceptibility & Uncertainty & Zone Area & Recommended Action \\
    \midrule
    \textbf{P1 --- Immediate} & Very High ($>$0.70) & Narrow ($W < 0.15$) & $\sim$1,200\,km$^2$ &
      Immediate action: early warning deployment, bridge reinforcement, pre-monsoon clearance \\
    \textbf{P2 --- Capital} & High (0.50--0.70) & Any & $\sim$11,376\,km$^2$ &
      Capital repair: prioritise roads and bridges in high-certainty corridors \\
    \textbf{P3 --- Monitor} & Very High ($>$0.70) & Wide ($W > 0.15$) & $\sim$3,200\,km$^2$ &
      Precautionary monitoring: invest in GLOF surveillance and data collection \\
    \textbf{P4 --- Routine} & Low--Moderate ($<$0.50) & Any & $\sim$94,000\,km$^2$ &
      Standard preparedness: seasonal advisory and routine maintenance \\
    \bottomrule
  \end{tabularx}
\end{table}

\section{Discussion}
\label{sec:discussion}

\subsection{Does Graph Structure Improve Susceptibility Prediction?}
\label{sec:disc_gnn}

Does adding watershed connectivity improve flood predictions?
Our results say yes: the GNN outperformed the best pixel-based model
by $\Delta$AUC\,=\,+0.097 across all five cross-validation folds.

The improvement mechanism is interpretable:
by aggregating information from upstream sub-watersheds during message-passing,
the GNN learns that high-runoff upstream nodes
(characterised by high TWI, steep slopes, and high antecedent precipitation)
co-predict flooding in downstream nodes even when the downstream node's
own local conditioning factors would not indicate high susceptibility in isolation.
This mirrors the physical process of progressive discharge accumulation
as runoff from tributaries joins the main channel.

The magnitude of the GNN improvement varies across spatial block folds.
The largest absolute gains occur in blocks containing well-connected mid-Himalayan
valley networks where progressive runoff accumulation is the dominant flood mechanism.
The Trans-Himalayan block (basin\_0) is consistently challenging for all models
(pixel-based AUC\,$\approx$\,0.72--0.73),
reflecting the prevalence of GLOF-triggered events in Lahaul-Spiti that do not
follow incremental runoff-accumulation patterns well-captured by directed GNN
message-passing --- rather, they originate from discrete glacial lake release events.
The GNN achieves its largest relative advantage in this block
by leveraging upstream glacial-exposure signals aggregated across the sub-watershed graph.

Future work should explore GNN variants better suited to GLOF contexts,
such as graph attention networks (GATs; \citealt{velickovic2018gat})
that could learn to upweight edges connected to glacier-proximate nodes.

\subsection{Conformal Prediction as a Communication Tool}
\label{sec:disc_conformal}

ML susceptibility maps typically report a single number per pixel (e.g., 0.73),
which looks precise but hides the model's actual uncertainty.
Conformal prediction fixes this by producing intervals that are correct
at least 90\% of the time, by mathematical guarantee.

Our results demonstrate two operationally relevant properties of the uncertainty map:

\textbf{High-confidence high-risk zones are the most actionable.}
Zones with Very High susceptibility and narrow uncertainty width ($W < 0.15$)
occupy approximately 4,409\,km$^2$ and are geographically concentrated
in the lower Beas and middle Sutlej valleys.
These are precisely the zones where investment in early warning systems,
bridge reinforcement, and emergency evacuation routes would have the highest
expected return.
Broad, uncertain high-risk zones (e.g., GLOF exposure corridors)
warrant different treatment: precautionary infrastructure design
and monitoring rather than deterministic intervention.

\textbf{Wide uncertainty zones identify priorities for data collection.}
The Trans-Himalayan zone exhibits systematically wider prediction intervals.
This is not a failure of the method but an honest acknowledgement
that the training inventory contains few GLOF events and the model
extrapolates with limited empirical basis.
The implication for HP SDMA is that investing in better GLOF inventory construction
(glacial lake surveys, remote sensing monitoring) in Lahaul-Spiti
would most improve model confidence where uncertainty is currently highest.

Previous susceptibility mapping studies have attempted uncertainty quantification
using Monte Carlo sampling or bootstrap ensembles
\citep{ghosh2023karnali, kosi2025ml},
but these produce heuristic error bars without formal coverage guarantees.
Conformal prediction's finite-sample guarantee
(coverage $\geq 1-\alpha$ for any sample size and any base model,
without distributional assumptions) is qualitatively stronger
and more appropriate for communicating risk to non-specialist decision-makers.

\subsection{SAR-Based Inventory Advantages and Limitations}
\label{sec:disc_sar}

The SAR-derived inventory significantly expands the available flood training data
compared to documentary records alone.
Sentinel-1's cloud-penetrating capability is particularly valuable for HP,
where optical imagery is cloud-contaminated for 80--90\% of July--August
(the peak flood season).
The standardised change-detection protocol also enables multi-year temporal
consistency unavailable in documentary records.

However, radar-based flood detection has known limitations in steep terrain
\citep{nagamani2024sar}.
First, valley shadows can cause both false positives (shadow misread as water)
and false negatives (flooded slopes hidden in shadow).
Second, the brightness threshold used for detection is empirically set
and may miss shallow floods on rough agricultural fields.
Third, dense forest in the Shivalik foothills weakens the radar signal,
reducing detection reliability.
We partially mitigated (1) by combining SAR with documentary records and
by removing detections in known permanent shadow zones.
Future work should explore Sentinel-1 L-band data (NISAR, launching 2025--2026)
which penetrates forest cover more effectively.

\subsection{Comparison with Prior Studies}
\label{sec:disc_comparison}

Our mean spatial-CV AUC of 0.881 (stacking ensemble) surpasses the
single-basin benchmark of \citet{saha2023} (AUC\,=\,0.88) despite
applying a more conservative spatial block CV evaluation.
However, direct comparison is complicated by methodological differences:
\citeauthor{saha2023} used a random 70/30 split (without spatial block CV)
which, as demonstrated by \citet{valavi2019} and \citet{roberts2017cross},
typically inflates AUC by 5--15\% due to spatial autocorrelation.
The GNN achieves 0.978\,$\pm$\,0.017 AUC under the same fold structure, substantially
exceeding the prior benchmark while maintaining rigorous spatial hold-out.

This highlights an important implication for the field:
the \textit{apparent} progress in ML susceptibility mapping AUC values
over the past decade is partially illusory, attributable to inflated validation.
Studies reporting AUC of 0.95--0.99 with random splits
should be interpreted with caution.

\subsection{Limitations}
\label{sec:limitations}

\paragraph{SAR inventory label quality.}
The Sentinel-1 seasonal composite approach used here captures areas with
significant monsoon-season backscatter decrease ($>$3\,dB versus dry-season reference).
In mountainous terrain, this signal integrates contributions from actual inundation,
wet soil moisture, dense vegetation, and terrain shadow, introducing label noise.
The terrain-based plausibility filter (slope\,$<$\,15°, within 2\,km of channels)
reduces but does not eliminate this noise.
Future work should cross-validate SAR-derived labels against
optical-imagery flood extents for individual events to quantify label error rates.
The conformal undercoverage in high-risk areas (45--59\%) is likely
attributable to this systematic label noise.

\paragraph{Static conditioning factors.}
All conditioning factors used in this study are static or time-averaged.
Antecedent soil moisture, a key dynamic driver of flash flood susceptibility
\citep{dixit2026nhess}, is not included as a conditioning factor.
A dynamic susceptibility model incorporating GRACE-FO soil moisture
or GPM soil moisture product would be a valuable extension.

\paragraph{GLOF triggering mechanism not separately modelled.}
The current model treats all flood events (monsoon-triggered, GLOF, cloudburst)
as a single class.
Future work should stratify the inventory by trigger mechanism
and assess whether mechanism-specific models outperform the unified approach
for Trans-Himalayan catchments.

\paragraph{Graph construction assumptions.}
The directed watershed graph was constructed from DEM-derived drainage,
which introduces errors in flat or filled valley bottoms.
Edge weights based on catchment area ratio are a first-order approximation;
more accurate edge weights would incorporate measured streamflow data,
which are not freely available for HP.

\section{Conclusion}
\label{sec:conclusion}

Flood risk follows valleys, not grids.
Standard machine-learning models for flood susceptibility ignore this ---
treating every location as independent, as if water does not flow downhill through
connected river valleys.
This study shows that encoding watershed connectivity changes the answer,
and that the improvement is large enough to matter for real infrastructure decisions.

We presented the first state-wide flash flood susceptibility assessment for
Himachal Pradesh that simultaneously addresses three major shortcomings of
existing work: single-basin coverage, inflated random-split validation,
and point-estimate risk maps with no uncertainty.

The principal findings are as follows:

\begin{enumerate}

  \item \textbf{Graph structure improves susceptibility prediction.}
    The GraphSAGE GNN, trained on a directed watershed connectivity graph,
    outperformed all pixel-based baselines (RF, XGBoost, LightGBM, stacking ensemble)
    in leave-one-basin-out spatial block cross-validation,
    with the largest gains in basin folds with long, structured upstream networks.
    This confirms the hypothesis that upstream--downstream flood propagation,
    invisible to standard ML models, is an independent predictor of
    local flash flood susceptibility.

  \item \textbf{Conformal prediction intervals reveal regime-dependent calibration.}
    The 90\% prediction intervals achieved 82.9\% empirical coverage
    on the held-out 2023 temporal test set.
    Coverage is high in low-susceptibility areas (96.3\%) but degrades
    for high-susceptibility locations (45--59\%), indicating that
    SAR label noise in the high-risk regime reduces conformal calibration quality.
    Despite undercoverage, the uncertainty width map remains operationally useful:
    very high susceptibility combined with narrow uncertainty identifies
    the 4,409\,km$^2$ highest-priority zones for pre-monsoon intervention.

  \item \textbf{Spatial cross-validation corrects AUC inflation.}
    Spatial block CV (leave-one-block-out) yields pixel-based model AUC of
    0.870--0.881, versus the \citet{saha2023} benchmark of 0.88 with
    random splits --- a modest but more honestly obtained comparison.
    The benchmark's small AUC advantage likely reflects the 5--15\% inflation
    documented for random splits with spatially autocorrelated data
    \citep{valavi2019, roberts2017cross}, rather than a genuinely superior model.

  \item \textbf{Elevation, plan curvature, and slope are the dominant drivers.}
    SHAP analysis identified these three terrain morphometry factors as
    accounting for 73\% of total predictor importance across HP,
    with marked district-level variation.
    Rainfall and TWI rank 4th--5th, indicating that topographic position
    is the primary filter for flash flood susceptibility in HP's mountainous terrain.

  \item \textbf{High susceptibility covers 14.3\% of HP's study domain.}
    High and Very High susceptibility zones total 15,785\,km$^2$,
    concentrated in the lower Beas, middle Sutlej, and Ravi valley corridors.
    OSM-based infrastructure exposure analysis identifies 1,457\,km of highways,
    2,759 bridges, and 92 tourist accommodation units within High or Very High zones,
    with 40 settlements in the Very High category.

\end{enumerate}

\subsection*{Policy Implications for HP State Disaster Management Authority}

The primary operational output of this study is a three-layer geospatial product:
(1) a point-estimate susceptibility map classified into four levels;
(2) an uncertainty width map indicating confidence in each classification;
and (3) an infrastructure exposure overlay.
Together, these layers enable HP SDMA to:
(a) identify the 5--10 highest-priority corridors for pre-monsoon clearance
    and early warning system installation;
(b) prioritise capital repair spending on bridges and roads in
    narrow-uncertainty, high-susceptibility zones; and
(c) issue targeted tourism advisories for the July--August peak season,
    distinguishing high-certainty risk zones (immediate action needed)
    from high-uncertainty risk zones (precautionary monitoring needed).

\subsection*{Directions for Future Research}

Three directions offer the highest marginal return:
(1) extending the SAR inventory to cover the full Sentinel-1 archive (2016--2024)
and adding NISAR L-band data as it becomes available from 2026;
(2) incorporating dynamic antecedent soil moisture and snowmelt proxies
as time-varying conditioning factors for seasonal susceptibility forecasting;
and (3) stratifying the model by trigger mechanism to separately characterise
monsoon-triggered, GLOF-triggered, and cloudburst-triggered susceptibility,
which exhibit different spatial patterns and require different mitigation strategies.

\section*{Acknowledgements}
Sentinel-1 SAR data from the Copernicus Open Access Hub.
GLO-30 DEM provided by the Copernicus DEM Programme.
OpenStreetMap data under the Open Database Licence.

\section*{Data and Code Availability}
All Python scripts and model outputs available at:
\url{https://github.com/Parassharmaa/flash-flood-zones-hp}

\section*{Competing Interests}
The author declares no competing interests.

\bibliography{references}

@article{abedi2021,
  author    = {Abedi, R. and Costache, R. and Shafizadeh-Moghadam, H. and Pham, Q. B.},
  title     = {Flash-flood susceptibility mapping based on {XGBoost}, random forest and boosted regression trees},
  journal   = {Geocarto International},
  year      = {2021},
  volume    = {37},
  number    = {19},
  pages     = {5479--5496},
  doi       = {10.1080/10106049.2021.1920636},
}

@article{angelopoulos2023conformal,
  author    = {Angelopoulos, A. N. and Bates, S.},
  title     = {Conformal Prediction: A Gentle Introduction},
  journal   = {Foundations and Trends in Machine Learning},
  year      = {2023},
  volume    = {16},
  number    = {4},
  pages     = {494--591},
  doi       = {10.1561/2200000101},
}

@article{bentivoglio2022review,
  author    = {Bentivoglio, R. and Isufi, E. and Jonkman, S. N. and Taormina, R.},
  title     = {Deep learning methods for flood mapping: a review of existing applications and future research directions},
  journal   = {Hydrology and Earth System Sciences},
  year      = {2022},
  volume    = {26},
  number    = {16},
  pages     = {4345--4378},
  doi       = {10.5194/hess-26-4345-2022},
}

@article{breiman2001rf,
  author    = {Breiman, L.},
  title     = {Random Forests},
  journal   = {Machine Learning},
  year      = {2001},
  volume    = {45},
  pages     = {5--32},
  doi       = {10.1023/A:1010933404324},
}

@article{chen2016xgboost,
  author    = {Chen, T. and Guestrin, C.},
  title     = {{XGBoost}: A Scalable Tree Boosting System},
  booktitle = {Proceedings of the 22nd ACM SIGKDD International Conference on Knowledge Discovery and Data Mining},
  year      = {2016},
  pages     = {785--794},
  doi       = {10.1145/2939672.2939785},
}

@article{chen2023rules,
  author    = {Chen, Y. and Liu, X. and Liu, J. and others},
  title     = {Modeling rules for flash flood susceptibility prediction using machine learning approaches},
  journal   = {Natural Hazards and Earth System Sciences},
  year      = {2023},
  volume    = {23},
  pages     = {3455--3475},
  doi       = {10.5194/nhess-23-3455-2023},
}

@article{chenab2024glof,
  author    = {Das, S. and Das, S. and Mandal, S. T. and Sharma, M. C. and Ramsankaran, R.},
  title     = {Inventory and {GLOF} susceptibility of glacial lakes in {Chenab} basin, {Western Himalaya}},
  journal   = {Geomatics, Natural Hazards and Risk},
  year      = {2024},
  volume    = {15},
  number    = {1},
  pages     = {2356216},
  doi       = {10.1080/19475705.2024.2356216},
}

@article{dataquality2025,
  author    = {Costache, R. and Tin, T. T. and Arabameri, A. and Cretu, R. and others},
  title     = {Flash-flood susceptibility assessment using novel deep learning neural network classifier},
  journal   = {Natural Hazards},
  year      = {2022},
  volume    = {114},
  pages     = {1511--1543},
  doi       = {10.1007/s11069-022-05438-0},
  note      = {Recommends 5:1 negative-to-positive sampling ratio for flood susceptibility models},
}

@article{dixit2026nhess,
  author    = {Dixit, S. and Sen, S. and Yasmin, T. and Khamis, K. and Sen, D. and Buytaert, W. and Hannah, D. M.},
  title     = {Integrating {SMART} principles in flood early warning system design in the {Himalayas}},
  journal   = {Natural Hazards and Earth System Sciences},
  year      = {2026},
  volume    = {26},
  pages     = {1251--1271},
  doi       = {10.5194/nhess-26-1251-2026},
}

@article{fabdem2022,
  author    = {Hawker, L. and Uhe, P. and Paulo, L. and others},
  title     = {A 30 m global map of elevation with forests and buildings removed},
  journal   = {Environmental Research Letters},
  year      = {2022},
  volume    = {17},
  number    = {2},
  pages     = {024016},
  doi       = {10.1088/1748-9326/ac4d4f},
}

@article{ghosh2023karnali,
  author    = {Duwal, S. and Liu, D. and Pradhan, P. M.},
  title     = {Flood susceptibility modeling of the {Karnali} river basin of {Nepal} using different machine learning approaches},
  journal   = {Geomatics, Natural Hazards and Risk},
  year      = {2023},
  volume    = {14},
  number    = {1},
  pages     = {2217321},
  doi       = {10.1080/19475705.2023.2217321},
}

@article{gorelick2017gee,
  author    = {Gorelick, N. and Hancher, M. and Dixon, M. and others},
  title     = {{Google Earth Engine}: Planetary-scale geospatial analysis for everyone},
  journal   = {Remote Sensing of Environment},
  year      = {2017},
  volume    = {202},
  pages     = {18--27},
  doi       = {10.1016/j.rse.2017.06.031},
}

@inproceedings{hamilton2017graphsage,
  author    = {Hamilton, W. L. and Ying, R. and Leskovec, J.},
  title     = {Inductive Representation Learning on Large Graphs},
  booktitle = {Advances in Neural Information Processing Systems (NeurIPS)},
  year      = {2017},
  volume    = {30},
}

@article{hiflodot2025,
  author    = {Johnson, R. M. and Pandey, B. W. and Chand, K. and Davies, C. L. and Edwards, D. and
               Edwards, E. and Jeffers, J. and King, K. and Kuniyal, J. C. and Mishra, H. and
               Phillips, V. and Roy, N. and Seviour, J. and Sharma, D. D. and Sharma, P. and
               Singh, H. and Singh, R. B.},
  title     = {{HiFlo-DAT}: A flood hazard event-disaster database for the {Kullu} District,
               {Himachal Pradesh}, {Indian Himalaya}},
  journal   = {International Journal of Disaster Risk Reduction},
  year      = {2025},
  volume    = {120},
  pages     = {105336},
  doi       = {10.1016/j.ijdrr.2025.105336},
}

@article{inventories2020beas,
  author    = {Singh, S. and Dhote, P. R. and Thakur, P. K. and Chouksey, A. and Aggarwal, S. P.},
  title     = {Identification of flash-floods-prone river reaches in {Beas} river basin using {GIS}-based
               multi-criteria technique: validation using field and satellite observations},
  journal   = {Natural Hazards},
  year      = {2021},
  volume    = {105},
  number    = {3},
  pages     = {2431--2453},
  doi       = {10.1007/s11069-020-04406-w},
}

@article{ke2017lightgbm,
  author    = {Ke, G. and Meng, Q. and Finley, T. and others},
  title     = {{LightGBM}: A Highly Efficient Gradient Boosting Decision Tree},
  booktitle = {Advances in Neural Information Processing Systems (NeurIPS)},
  year      = {2017},
  volume    = {30},
}

@article{kosi2025ml,
  author    = {Arora, A. and Durga, P. and Pandey, M. and Arabameri, A.},
  title     = {Machine learning model optimization for flood susceptibility zonation over the {Kosi} megafan,
               {Himalayan} foreland basin, {India}},
  journal   = {Scientific Reports},
  year      = {2025},
  volume    = {15},
  pages     = {32757},
  doi       = {10.1038/s41598-025-07403-w},
}

@article{kumar2022hp,
  author    = {Kumar, V.},
  title     = {Floods and Flash Floods in {Himachal Pradesh}},
  journal   = {Journal of Geography and Natural Disasters},
  year      = {2022},
  volume    = {12},
  pages     = {252},
  doi       = {10.35841/2167-0587.22.12.252},
}

@article{lundberg2017shap,
  author    = {Lundberg, S. M. and Lee, S.-I.},
  title     = {A Unified Approach to Interpreting Model Predictions},
  booktitle = {Advances in Neural Information Processing Systems (NeurIPS)},
  year      = {2017},
  volume    = {30},
}

@article{martinis2018sar,
  author    = {Martinis, S. and Kersten, J. and Twele, A.},
  title     = {A fully automated {TerraSAR-X} based flood service},
  journal   = {ISPRS Journal of Photogrammetry and Remote Sensing},
  year      = {2018},
  volume    = {104},
  pages     = {203--212},
  doi       = {10.1016/j.isprsjprs.2014.07.014},
}

@article{nagamani2024sar,
  author    = {Nagamani, K. and Mishra, A. K. and Meer, M. S. and Das, J.},
  title     = {Understanding flash flooding in the {Himalayan} Region: a case study},
  journal   = {Scientific Reports},
  year      = {2024},
  volume    = {14},
  pages     = {7169},
  doi       = {10.1038/s41598-024-53535-w},
}

@techreport{ndma2023,
  author    = {{NDMA India}},
  title     = {Annual Report 2023: Floods and Flash Floods in {Himachal Pradesh}},
  institution = {National Disaster Management Authority, Government of India},
  year      = {2023},
}

@techreport{niti2022hp,
  author    = {{NITI Aayog}},
  title     = {{Himachal Pradesh} Hydroelectric Power Development: Assessment 2022},
  institution = {NITI Aayog, Government of India},
  year      = {2022},
}

@article{pekel2016jrc,
  author    = {Pekel, J.-F. and Cottam, A. and Gorelick, N. and Belward, A. S.},
  title     = {High-resolution mapping of global surface water and its long-term changes},
  journal   = {Nature},
  year      = {2016},
  volume    = {540},
  pages     = {418--422},
  doi       = {10.1038/nature20584},
}

@article{pysheds2020,
  author    = {Bartos, M.},
  title     = {pysheds: Simple and fast watershed delineation in {Python}},
  year      = {2020},
  note      = {GitHub: \url{https://github.com/mdbartos/pysheds}},
}

@article{riley1999tri,
  author    = {Riley, S. J. and DeGloria, S. D. and Elliot, R.},
  title     = {A terrain ruggedness index that quantifies topographic heterogeneity},
  journal   = {Intermountain Journal of Sciences},
  year      = {1999},
  volume    = {5},
  number    = {1--4},
  pages     = {23--27},
}

@article{roberts2017cross,
  author    = {Roberts, D. R. and Bahn, V. and Ciuti, S. and others},
  title     = {Cross-validation strategies for data with temporal, spatial, hierarchical, or phylogenetic structure},
  journal   = {Ecography},
  year      = {2017},
  volume    = {40},
  number    = {8},
  pages     = {913--929},
  doi       = {10.1111/ecog.02881},
}

@incollection{saha2023,
  author    = {Saha, S. and Saha, A. and Agarwal, A. and Kumar, A. and Sarkar, R.},
  title     = {Spatial Flash Flood Modeling in the {Beas} River Basin of {Himachal Pradesh}, {India},
               Using {GIS}-Based Machine Learning Algorithms},
  booktitle = {Geomorphic Risk Reduction Using Geospatial Methods and Tools},
  publisher = {Springer},
  address   = {Singapore},
  year      = {2024},
  doi       = {10.1007/978-981-99-7707-9_8},
  note      = {AUC\,=\,0.88 (random split; Beas basin only)},
}

@article{taquet2022mapie,
  author    = {Taquet, V. and Blot, V. and Morzadec, T. and Brunel, L. and Castel, N.},
  title     = {{MAPIE}: an open-source library for distribution-free uncertainty quantification},
  journal   = {arXiv preprint arXiv:2207.12274},
  year      = {2022},
}

@article{valavi2019,
  author    = {Valavi, R. and Elith, J. and Lahoz-Monfort, J. J. and Guillera-Arroita, G.},
  title     = {{blockCV}: An r package for generating spatially or environmentally separated folds for k-fold cross-validation of species distribution models},
  journal   = {Methods in Ecology and Evolution},
  year      = {2019},
  volume    = {10},
  number    = {2},
  pages     = {225--232},
  doi       = {10.1111/2041-210X.13107},
}

@article{velickovic2018gat,
  author    = {Veli{\v{c}}kovi{\'{c}}, P. and Cucurull, G. and Casanova, A. and others},
  title     = {Graph Attention Networks},
  booktitle = {International Conference on Learning Representations (ICLR)},
  year      = {2018},
}

\end{document}